\documentclass[journal]{IEEEtran}

\ifCLASSOPTIONcompsoc
  \usepackage[nocompress]{cite}
\else
  \usepackage{cite}
\fi

\usepackage{graphicx}
\usepackage{subfigure}
\usepackage{multirow}
\usepackage{amssymb}
\usepackage{amsmath}
\usepackage{booktabs}
\usepackage[ruled]{algorithm2e}
\usepackage{color}
\usepackage{ragged2e}
\usepackage{hyperref}
\usepackage{cite}

\ifCLASSINFOpdf
\else
\fi

\begin{document}
%

\title{Pruning Networks with Cross-Layer Ranking \& \textit{k}-Reciprocal Nearest Filters}

\author{Mingbao Lin,
        Liujuan Cao,
        Yuxin Zhang,
        Ling Shao, ~\IEEEmembership{Fellow,~IEEE},\\
        Chia-Wen Lin,~\IEEEmembership{Fellow, ~IEEE}
        and Rongrong Ji,~\IEEEmembership{Senior Member,~IEEE},
\IEEEcompsocitemizethanks{
\IEEEcompsocthanksitem M. Lin is with the Media Analytics and Computing Laboratory, Department of Artificial Intelligence, School of Informatics, Xiamen University, Xiamen 361005, China, also with the Youtu Laboratory, Tencent, Shanghai 200233, China.
\IEEEcompsocthanksitem L. Cao (Corresponding Author) is with the Fujian Key Laboratory of Sensing and Computing for Smart City, Computer Science Department, School of Informatics, Xiamen University, Xiamen 361005, China (e-mail: caoliujuan@xmu.edu.cn).
\IEEEcompsocthanksitem Y. Zhang is with the Media Analytics and Computing Laboratory, Department of Artificial Intelligence, School of Informatics, Xiamen University, Xiamen 361005, China.\protect
\IEEEcompsocthanksitem L. Shao is with the Inception Institute of Artificial Intelligence, Abu Dhabi, United Arab Emirates, and also with the Mohamed bin Zayed University of Artificial Intelligence, Abu Dhabi, United Arab Emirates.\protect
\IEEEcompsocthanksitem C.-W. Lin is with the Department of Electrical Engineering and the Institute of Communications Engineering, National Tsing Hua University, Hsinchu 30013, Taiwan.
\IEEEcompsocthanksitem R. Ji is with the Media Analytics and Computing Laboratory, Department of Artificial Intelligence, School of Informatics, Xiamen University, Xiamen 361005, China, also with the Institute of Artificial Intelligence, Xiamen University, Xiamen 361005, China.\protect
}
\thanks{Manuscript received April 19, 2005; revised August 26, 2015.}}

\markboth{IEEE Transactions on Neural Networks and Learning Systems Under Review}%
{Shell \MakeLowercase{\textit{et al.}}: Bare Demo of IEEEtran.cls for IEEE Journals}

\maketitle

\begin{abstract}
This paper focuses on filter-level network pruning. A novel pruning method, termed CLR-RNF, is proposed. We first reveal a ``long-tail'' pruning problem in magnitude-based weight pruning methods, and then propose a computation-aware measurement for individual weight importance, followed by a Cross-Layer Ranking (CLR) of weights to identify and remove the bottom-ranked weights. Consequently, the per-layer sparsity makes up of the pruned network structure in our filter pruning. Then, we introduce a recommendation-based filter selection scheme where each filter recommends a group of its closest filters. To pick the preserved filters from these recommended groups, we further devise a \textit{k}-Reciprocal Nearest Filter (RNF) selection scheme where the selected filters fall into the intersection of these recommended groups. Both our pruned network structure and the filter selection are non-learning processes, which thus significantly reduce the pruning complexity, and differentiate our method from existing works. We conduct image classification on CIFAR-10 and ImageNet to demonstrate the superiority of our CLR-RNF over the state-of-the-arts. For example, on CIFAR-10, CLR-RNF removes 74.1\% FLOPs and 95.0\% parameters from VGGNet-16 with even 0.3\% accuracy improvements. On ImageNet, it removes 70.2\% FLOPs and 64.8\% parameters from ResNet-50 with only 1.7\% top-5 accuracy drops. Our project is at \url{https://github.com/lmbxmu/CLR-RNF}.

\end{abstract}


\maketitle


\IEEEpeerreviewmaketitle

\section{Introduction}\label{introduction}

\IEEEPARstart{T}{hough} deep convolutional neural networks (CNNs) are prevailing, it comes at the cost of huge computational burden and large power consumption, which poses a great challenge for real-time deployments on resource-limited devices such as cell phones and Internet-of-Things (IoT) devices. To address this problem, model compression has become an active research  topic, which aims to reduce the model redundancy with a comparable or even better performance  in comparison with the full model, such that the compressed model can be easily run on resource-limited devices.

General methods for reducing the model size can be roughly categorized into five groups: (1) Low-bit quantization aims to compress a pre-trained model by reducing the number of bits used to represent the weight parameters of the pre-trained models~\cite{Vanhoucke2011improving,Zhou2017incremental,lin2020rotated}. (2) Compact networks such as ShuffleNets~\cite{zhang2018shufflenet,ma2018shufflenet}, MobileNets~\cite{howard2017mobilenets,sandler2018mobilenetv2,howard2019searching} and GhostNet~\cite{han2020ghostnet}, directly design parameter-efficient neural network models. (3) Tensor factorization approximates the weight tensor with a series of low-rank matrices, which are then organized in a sum-product form~\cite{lin2018holistic,hayashi2019exploring}. 
(4) Network pruning removes a certain part of the network. According to the pruning granularity, existing methods include weight pruning~\cite{han2015learning,frankle2019lottery}, block pruning~\cite{zhou2021learning,choquette2021nvidia}, row/column pruning\cite{liu2020autocompress,zhang2018structadmm}, kernel pruning~\cite{mao2017exploring,xie2019exploring}, pattern pruning~\cite{elsen2020fast,kalchbrenner2018efficient}, filter pruning\cite{luo2018thinet,ding2019centripetal}, \emph{etc}.


%
In this paper, we focus on filter pruning for efficient image classification, which has received ever-increasing focus due to the following advantages:
1) The pruned model is structured, which can be well supported by regular hardware and off-the-shelf basic linear algebra subprograms (BLAS) library.
2) The storage usage and computational cost are significantly reduced in online inference.
3) It can be further combined with other compression methods, such as network quantization, tensor factorization, and weight pruning, to achieve a deeper compression and acceleration.
Despite the extensive progress~\cite{he2017channel,huang2018data,liu2019metapruning,lin2020hrank,li2020eagleeye} made in the literature, two essential issues remain as open problems in the filter pruning, \emph{i.e.}, the pruned network structure and the filter importance measurement.

As the first issue, the pruned network structure is related to the per-layer pruning rate. Setting these pruning rates for different layers has shown to significantly affect the final performance~\cite{liu2019rethinking,liu2019metapruning,lin2020channel}. To this end, existing methods resort to a series of complex learning steps, many of which focus on training from scratch with additional sparsity constraints. For instance, methods in~\cite{liu2017learning,zhao2019variational} employ joint-retraining with sparse requirements on the scaling factors of batch normalization layers, and the pruning rate in each layer relies on a given threshold. Huang \emph{et al}.~\cite{huang2018data} proposed to train CNNs with the 0-1 mask on each filter and the percentage of 1s in each layer makes up of the pruned network structure. The method in~\cite{luo2020autopruner} takes previous activation responses as inputs and generates a binary index code for pruning. Similar to~\cite{huang2018data}, the pruned network structure consists of the ratio of trained non-zero indexes. Dynamic pruning~\cite{lin2020Dynamic} incorporates a feedback scheme to reactivate the pruned filters, which thus achieves dynamic allocation of the sparsity in each layer. Another group~\cite{li2017pruning,lin2020hrank} requires human experts to designate the layer-wise pruning strategy, which is simple but quantitatively suboptimal. More recent works~\cite{liu2019metapruning,dong2019network,yu2019autoslim,lin2020channel} focus on search-based strategies, typically through network architecture search~\cite{dong2019network}, one-shot architecture search~\cite{yu2019autoslim}, or heuristic-based search algorithms such as evolutionary algorithm~\cite{liu2019metapruning} and artificial bee colony~\cite{lin2020channel}. Although search-based methods generally result in a better network structure, their search progress is extremely time-consuming.

As the second issues, the filter importance measurement identifies which filters in the pre-trained model should be preserved and inherited to initialize the pruned network structure. Existing works focus on measuring the individual filter importance. To this end, many of them resorts to preserving the most ``important'' filters by a certain criterion to estimate the filter importance, such as magnitude-based~\cite{liu2017learning}, zero percentage of output activation~\cite{hu2016network}, rank of feature map~\cite{lin2020hrank}. However, the methods in~\cite{hu2016network,lin2020hrank} are data-driven and add complexity in evaluation, and the method in~\cite{liu2017learning} is more effective in weight pruning~\cite{li2017pruning,frankle2019lottery} rather than filter pruning as demonstrated in~\cite{ye2018rethinking}. Besides, these methods usually require layer-wise fine-tuning to improve inference accuracy, which is also time-consuming. Training-from-scratch methods~\cite{liu2017learning,zhao2019variational,huang2018data,luo2020autopruner,lin2020Dynamic} preserve the weights of non-zero masked filters or filters with fewer sparse factors for the follow-up fine-tuning. The methods in~\cite{dong2019network,liu2019metapruning,lin2020channel} adopt a random measurement to assign filter weights with random Gaussian distribution, or randomly pick up some of the pre-trained filter weights. Besides, methods in~\cite{liu2019metapruning,yu2019autoslim} also require to train a large auxiliary network to predict the weights of potential pruned network structure, making the pruning more complex.

\begin{figure}[!t]
\begin{center}
\includegraphics[height=0.45\linewidth]{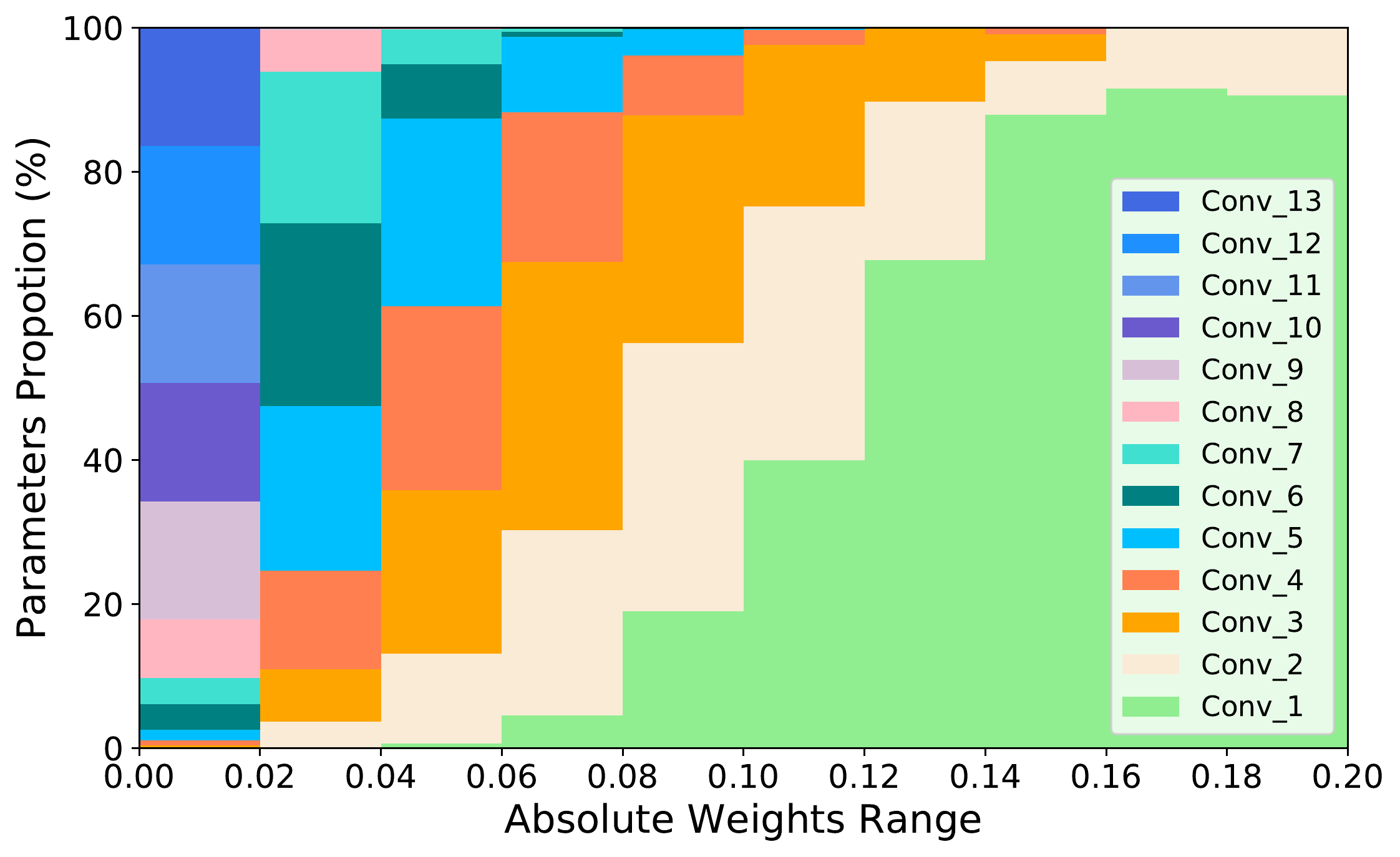}
\end{center}
\vspace{-1em}
\caption{\label{longtail}Illustration of the ``long-tail'' pruning problem in the magnitude-based weight pruning (VGGNet-16). The x-axis denotes different ranges of weight magnitude and the y-axis denotes the percentage of per-layer weights that fall into that interval. As can be seen, the top-layer weights are concentrated  in an interval of  small-magnitude range, while the bottom-layer weights span over a much larger magnitude range. Thus, much more top-layer weights tend to be removed in the magnitude-based weight pruning. 
\vspace{-1.5em}
}
\end{figure}

In this paper, we propose a novel pruning method, termed CLR-RNF, which consists of two components of CLR and RNF to respectively solve the above two problems. The former aims to efficiently find the optimal pruned network structure and the latter targets to select a subgroup of important filters to initialize the pruned network structure such that the pruned model performance can be effectively recovered. 
To find the optimal pruned network structure, we adopt the effective magnitude-based criterion in weight pruning~\cite{li2017pruning,frankle2019lottery} and introduce a cross-layer ranking (CLR) of weights. As a result, the pruned network structure with our filter pruning scheme also benefits from the per-layer sparsity employed in weight pruning. For the first time, we reveal the ``long-tail'' pruning problem in the magnitude-based weight pruning as illustrated in Fig.\,\ref{longtail}, and  propose a computation-aware measurement of weight importance to effectively address the inefficiency in network pruning caused by the long-tail. 
To select a subgroup of important filters, instead of selecting filters based on their individual importance, we prefer to measuring the collective importance of a filter group for selection, which is based on our insight that per-layer filters are involved in a coalition to achieve a desired performance. Specifically, each filter in the pre-trained model would recommend a group of its closest filters which have a high potential to be inherited by the pruned model. Correspondingly, a \textit{k}-reciprocal nearest filter (RNF) selection is proposed to pick up filters that fall into the intersection of all the recommended groups as the final inherited filters. Both our pruned network structure and filter selection are non-learning, which thus greatly simplifies the complexity in filter pruning. 

To sum up, the main contributions of this work include:

\begin{itemize}
\item For the first time, we reveal the ``long-tail'' pruning problem in the magnitude-based weight pruning which degrades the efficacy of a pruned network, and propose  a new computation-aware measurement to effectively address the problem.

\item We propose to treat the per-layer sparsity in the cross-layer ranking of weight pruning as the per-layer pruning rate for filter pruning. To the best of our knowledge, this is the first work that utilizes the linkage between filter pruning and weight pruning.

\item We propose a novel recommendation-based filter selection scheme based on the \textit{k}-reciprocal nearest neighbors recommended by individual filters in a layer. The method selects a group of filters by taking into account the overall collective importance of the filter group,  rather than the importance of individual filters. 
\end{itemize}

The rest of this paper is organized as follows: In Sec.\,\ref{related}, we discuss the related work. Details of our proposed CLR-RNF are elaborated in Sec.\,\ref{methodology}. Sec.\,\ref{experiment} presents the experimental results. Finally, we conclude this paper in Sec.\,\ref{conclusion}.

\begin{figure*}[!t]
\begin{center}
\includegraphics[height=0.25\linewidth]{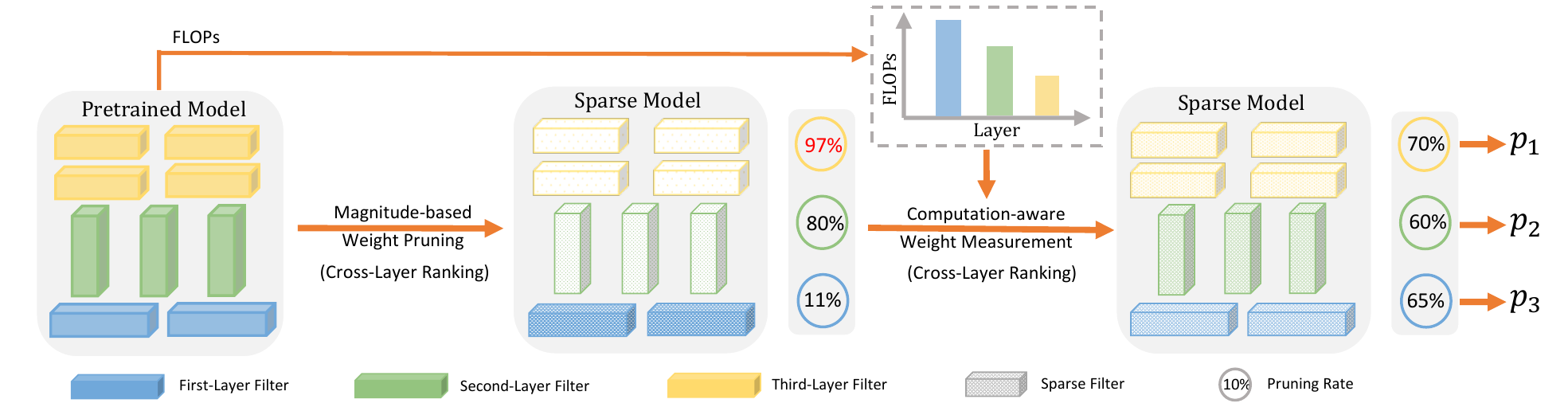}
\end{center}
\caption{\label{clr}Framework of our cross-layer ranking (CLR) for pruned network structure. More weight elements from the top layers are removed in the magnitude-based weight pruning due to their smaller weight values as shown in Fig.\,\ref{longtail}, which is termed as ``long-tail'' pruning problem in this paper. It results in fewer computation reductions since more FLOPs are accumulated in the bottom layers. Our computation-aware measurement integrates per-layer FLOPs into the weight importance estimate for a cross-layer weight ranking, which well balances the per-layer computation and per-layer sparsity, the latter of which makes up of our pruned network structure.
}
\end{figure*}

\section{Related Work}\label{related}
In what follows, we discuss the major topics that are the most related to our work.

\textbf{Weight Pruning}. In contrast to filter pruning, weight pruning pursues to remove individual neurons in the weight tensors of a neural network by a certain criterion or training technique, such as second-order Taylor expansion~\cite{lecun1990optimal}, second-order derivative~\cite{dong2017learning}, $\ell_2$-regularization~\cite{han2015learning}, global sparse momentum SGD~\cite{ding2019global}, and magnitude of weight value~\cite{zhu2018to,gal2019the,mostafa2019parameter,frankle2019lottery}. After removing the neurons, the weight tensors become highly sparse and the memory can be reduced by arranging the model in a sparse format. Specialized hardware and software are thus required to achieve practical speedups. Differently, we focus on filter pruning, but aim to make full use of the magnitude-based weight ranking to derive the pruned network structure.

\textbf{Neural Architecture Search}. Recently, neural architecture search (NAS) has attracted increasing attention~\cite{ren2020comprehensive}. It aims to design a network architecture in an automated way with as little human intervention as possible, typically through reinforcement learning~\cite{zoph2017neural}, evolutionary learning~\cite{real2017large}, differentiable search~\cite{liu2019darts} and so on. Similar to NAS, recent arts resort to search-based strategies for the pruned network structure~\cite{liu2019metapruning,lin2020channel}. Differently, the search space of NAS is broad (operations, filter number, and network depth, \emph{etc}.) and is defined distinctively across different works. On the contrary, filter pruning focuses on the decision of per-layer filter number to produce a subnet of a given network, which can be seen as a simplified version of architecture search.


\section{Methodology}\label{methodology}

\subsection{Preliminary}

Consider a pre-trained CNN with $L$ convolutional layers $\mathcal{C} = \{\mathcal{C}_1, \mathcal{C}_2, ..., \mathcal{C}_L\}$, whose kernels are $\mathbf{K} = \{ \mathbf{K}_1, \mathbf{K}_2, ..., \mathbf{K}_L \}$, where  $\mathcal{C}_i$ denotes the $i$-th convolutional layer with  kernel $\mathbf{K}_i = \{ \mathbf{k}_i^1, \mathbf{k}_i^2, ..., \mathbf{k}_i^{n_i} \}$ consisting of $n_i$ filters. The $j$-th filter of $\mathbf{K}_i$ can be represented by a three-way tensor $\mathbf{k}_i^j \in \mathbb{R}^{n_{i-1} \times h_i \times w_i}$, where $n_{i-1}$, $h_i$, and $w_i$ stand for the channel number, height, and width of the filter, respectively. As can be seen, the channel number in the $i$-th layer is equal to the filter number in the ($i-1$)-th layer. For ease of presentation, we reformat each filter with the shape of $\mathbf{k}_i^j \in \mathbb{R}^{n_{i-1} \cdot h_i \cdot w_i}$. We denote the $q$-th weight element in $\mathbf{k}_i^j$ as $(\mathbf{k}_i^j)_q \in \mathbf{k}_i^j$ and each weight in our setting would be assigned with an importance estimate denoted as $(\mathbf{\theta}_i^j)_q$.

Given a global pruning rate $p$, filter pruning aims to find and prune  redundant filters in each layer of a network to obtain a compressed representation of the pruned network $\bar{\mathbf{K}}_i = \{ \bar{\mathbf{k}}_i^1, \bar{\mathbf{k}}_i^2, ..., \bar{\mathbf{k}}_i^{\bar{n}_i} \} \subseteq \mathbf{K}_i$ with $\bar{\mathbf{k}}_i^j \in \mathbb{R}^{\bar{n}_{i-1} \cdot h_i \cdot w_i}$. By denoting the pruning rate in the $i$-th layer as $p_i$, we have $\bar{n}_i = \big\lceil (1 - p_i) \cdot n_i \big\rfloor$ where $\lceil \cdot \rfloor$ rounds its input to the nearest integer. $\bar{\mathbf{K}}_i$ is subsequently end-to-end fine-tuned to recover the accuracy performance.

As discussed in Sec.\,\ref{introduction}, the pruned network structure and the filter importance measurement are two important factors impacting the final pruning performance: the former reflected in the value of $p_i$ (or $\bar{n}_i$) and the latter reflected in the filters of $\mathbf{\bar{K}}_i$. To that effect, with $p$, prevalent methods resort to a series of complex learning in finding $p_i$ and focus on measuring the importance of individual filters to locate $\mathbf{\bar{K}}_i$. Instead, we aim to improve the filter pruning by proposing two non-learning components for finding a better pruned network structure and identifying a filter subset with a better collective importance.

\subsection{Cross-Layer Ranking}

Fig.\,\ref{clr} shows our policy for pruned network structure. Detailedly, our cross-layer ranking dates back to the weight pruning~\cite{han2015learning,zhu2018to,gal2019the,mostafa2019parameter,frankle2019lottery}, which directly measures the importance of each individual weight by its magnitude, \emph{i.e.},
\begin{equation}\label{magnitude}
(\mathbf{\theta}_i^j)_q = |(\mathbf{k}_i^j)_q|,
\end{equation}
where $| \cdot |$ returns the absolute value of its input. We then conduct a weight ranking across the whole network by the numerical order of $(\mathbf{\theta}_i^j)_q$. Given a global pruning rate $p$, it can be easily achieved by pruning out the lowest-rated weights. As a result, the weight pruning leads each filter $\mathbf{k}_i^j$ to a sparse $\hat{\mathbf{k}}_i^j$ whose elements are defined as
\begin{equation}\label{sparse}
(\hat{\mathbf{k}}_i^j)_q =
\begin{cases}
0 & (\theta_i^j)_q \; \text{is among the lowest-rated}, \\
(\mathbf{k}_i^j)_q & \text{otherwise}.
\end{cases}
\end{equation}

It has been demonstrated that more than 90\% of network parameters can be safely removed by weight pruning without compromising performance~\cite{lecun1990optimal,han2015learning,li2017pruning,frankle2019lottery} since weight pruning considers the ranking relationship across different layers thus the global redundancy can be tracked accurately. Besides global redundancy, we believe that per-layer sparsity in weight pruning provides useful information for determining the pruned network structure by filter pruning.

A straightforward method for designing the pruned network structure by filter pruning is to determine the per-layer pruning rate $p_i$ based on the per-layer sparsity after weight pruning as

\begin{equation}\label{pruning_rate}
p_i = \frac{\sum_j^{n_i}\sum_q^{n_{i-1} \cdot h_i \cdot w_i}\delta\big((\hat{\mathbf{k}}_i^j)_q \neq 0\big)}{n_i \cdot n_{i-1} \cdot h_i \cdot w_i},
\end{equation}
where $\delta(\cdot)$ is an indicator function, which returns 1 if the input is true, and 0 otherwise. 

%
\begin{figure}[!t]
\begin{center}
\begin{minipage}[t]{0.33\linewidth}
\centerline{
\subfigure[]{
\includegraphics[width=\linewidth]{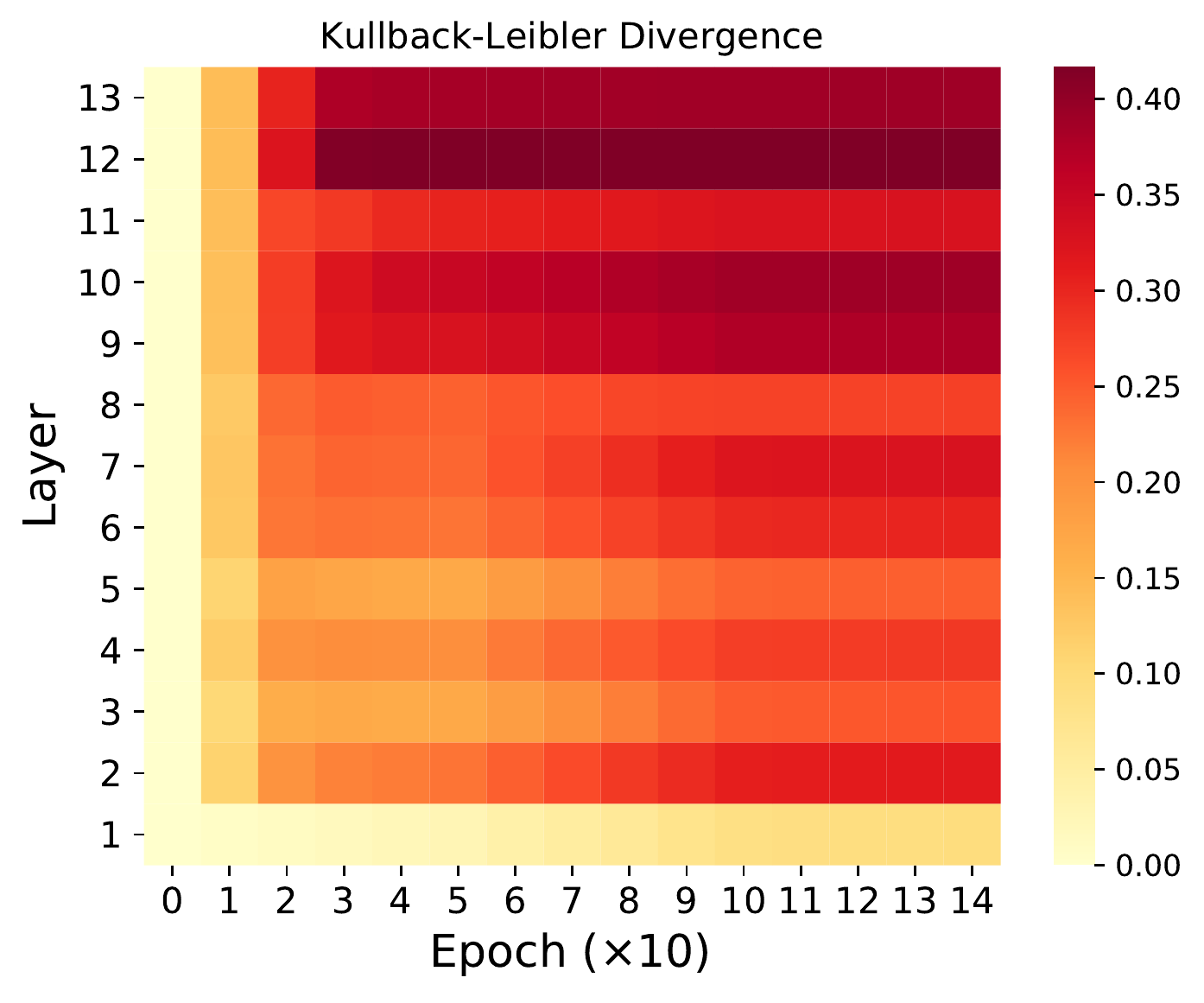}}
\subfigure[]{
\includegraphics[width=\linewidth]{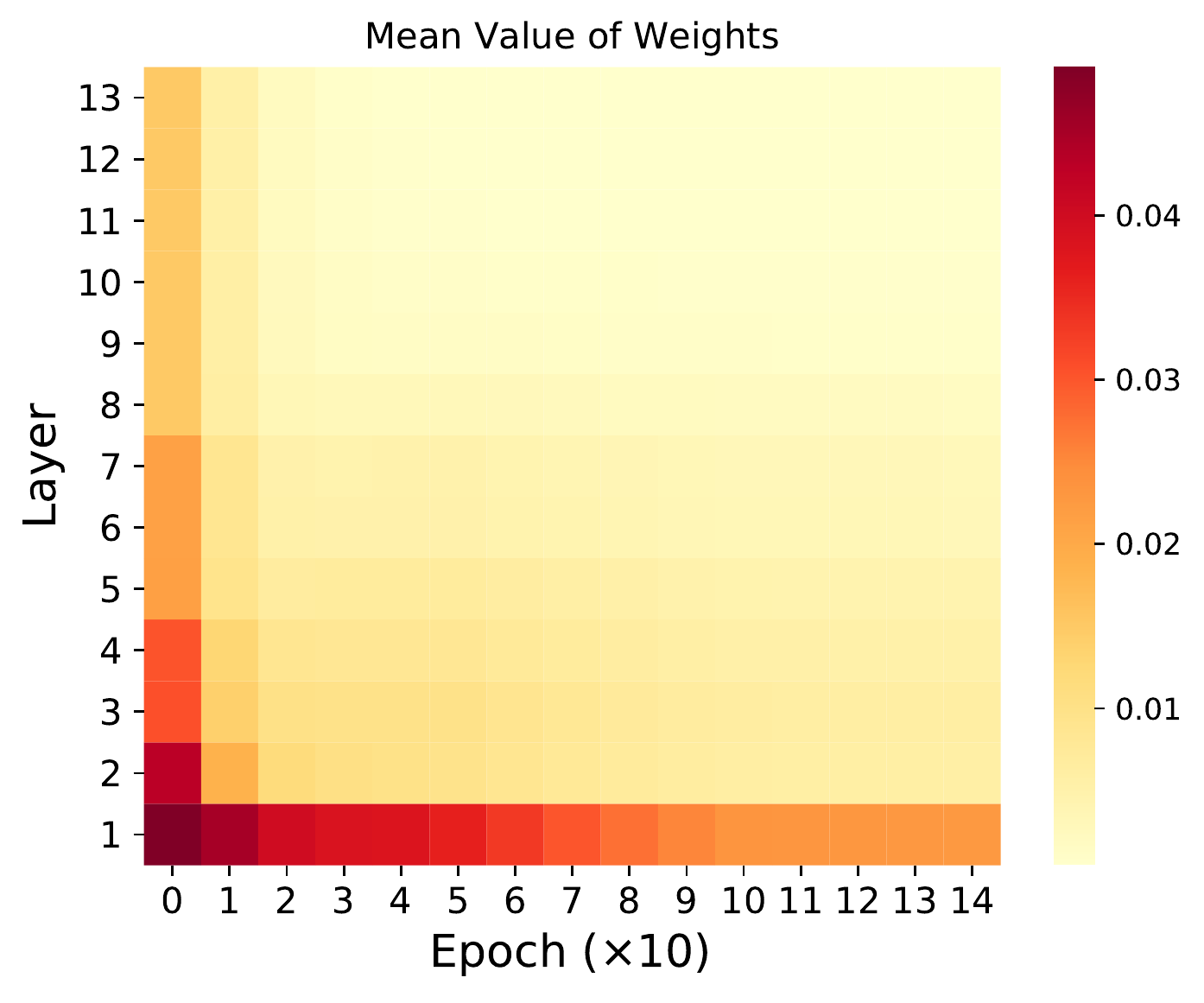}}
\subfigure[]{
\includegraphics[width=\linewidth]{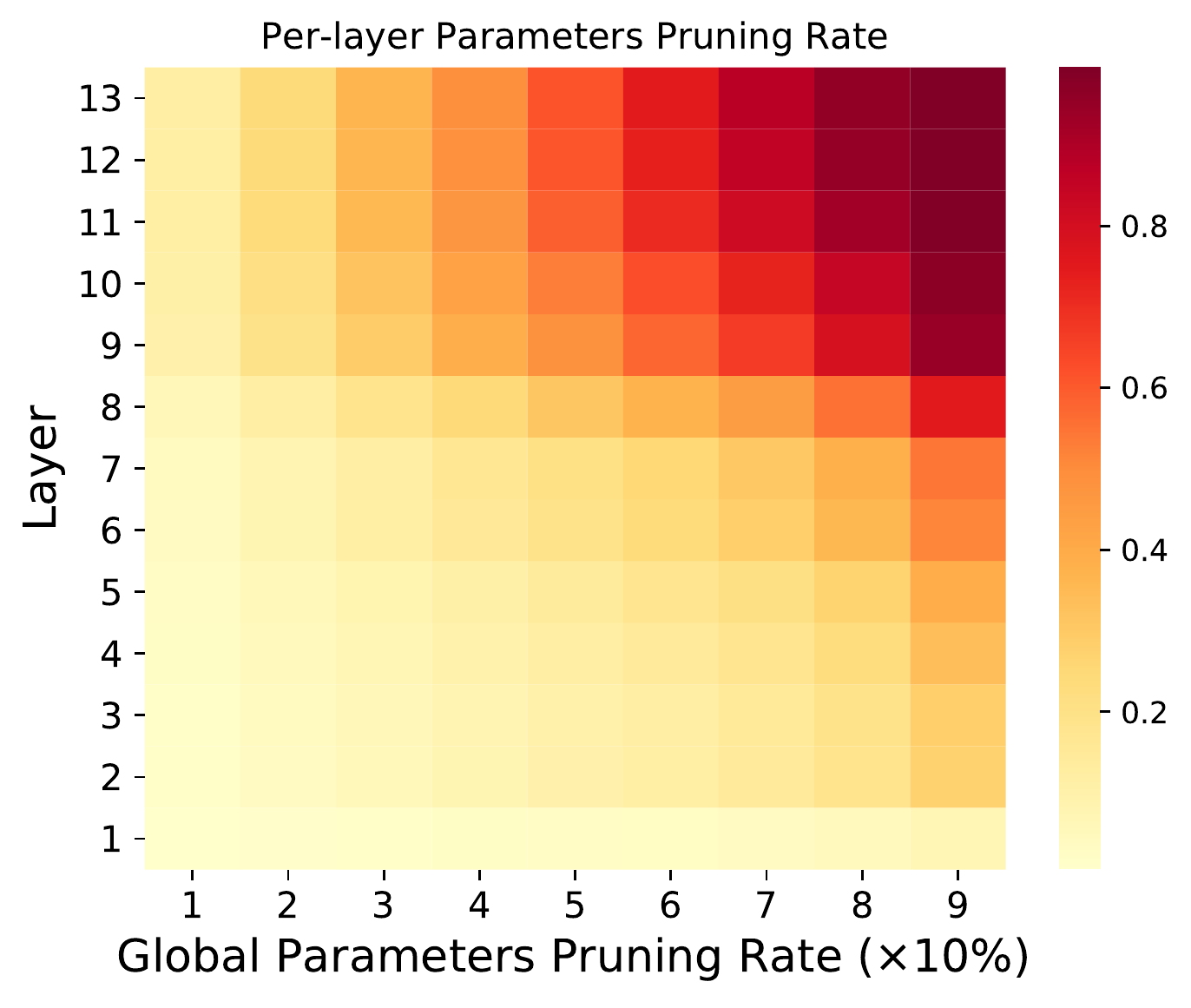}}
}
\end{minipage}
\end{center}
\caption{\label{visualization}Analysis on the ``long-tail'' pruning problem (VGGNet-16). (a) The KL-divergence between the initialization weights and weights trained at different training epochs. (b) The mean values of absolute weights at different training epochs. From (a) and (b), the top-layer weights change a lot and it leads to smaller weight values in the top network layers. Thus, the magnitude-based ranking will cause higher pruning rates in the top layers (c).}
\end{figure}

\begin{figure*}[!t]
\begin{center}
\includegraphics[height=0.25\linewidth]{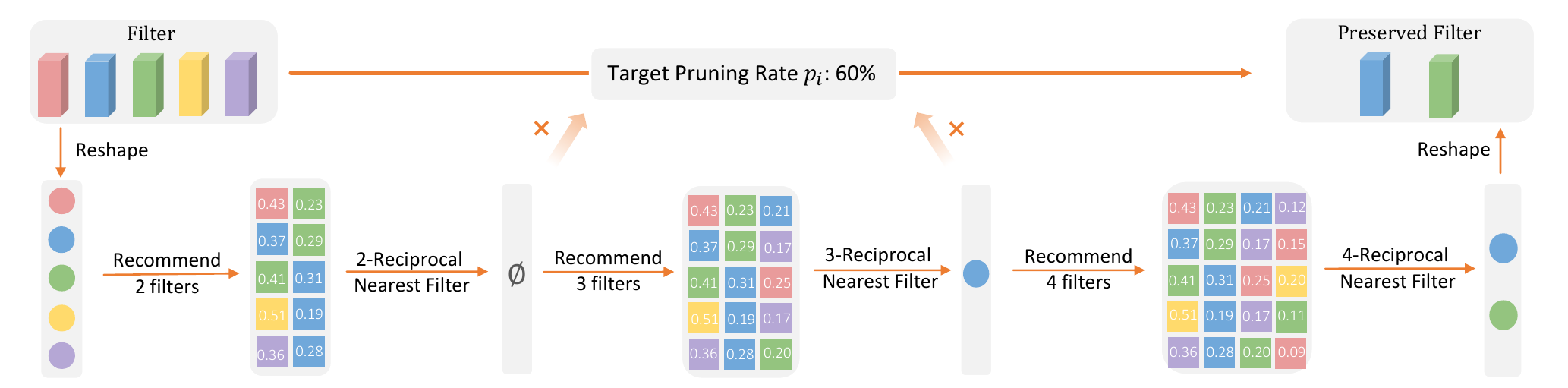}
\end{center}
\caption{\label{rnf}Framework of our $k$-reciprocal nearest filter (RNF) to determine a group of filter subsets for parameters transferred to the pruned network structure. The CLR in Fig.\,\ref{clr} feeds back the pruning rate $p_i$ ($p_i$ = 40\% in this illustration) and we can obtain the number of preserved filters $\bar{n}_i$ ($\bar{n}_i$ = 2 in this illustration). We first construct the similarity matrix to measure the closeness among the pre-trained filters. Our filter selection is recommendation-based, where each filter would recommend a group of filter subsets with higher closeness. Our $k$-reciprocal nearest filter picks up the intersection of all recommendations as the final selected filters. Starting with $k = \bar{n}_i$, we set $k = k + 1$ repeatedly until the intersection size satisfies $\bar{n}_i$.
}
\end{figure*}

The definition of $p_i$ in Eq.\,(\ref{pruning_rate}) equalizes the sparsity of the $i$-th layer in the weight pruning. However, the weight importance in Eq.\,(\ref{magnitude}) is closely related to the network compression while ignoring the FLOPs. in the network, which is directly related to the acceleration. We empirically observe that the magnitudes of weights in the top layers are usually smaller than those in the bottom layers as shown in Fig.\,\ref{longtail}. Consequently, the weight pruning~\cite{han2015learning,zhu2018to,gal2019the,mostafa2019parameter,frankle2019lottery} tends to remove more individual weights in the top layers, known as ``long-tail'' pruning.

To make an in-depth analysis, in Fig.\,\ref{visualization}(a), we show the Kull-Leibler divergence between the initial weights and the weights derived in different training stages. As can be observed, the trained weights in the top layers show a significant difference from the initial weights while the bottom-layer weights are nearly unchanged. Fig.\,\ref{visualization}(b) shows the per-layer mean of weight magnitudes in different training epochs and it is clear that the great changes in Fig.\,\ref{visualization}(a) result in smaller weight values in the top layers. We relate this phenomenon to the problem of ``gradient vanishing'' in network learning, \emph{i.e.}, the bottom-layer gradient is vanishingly small thus the filter weights keep unchanged during training. As a result, a large portion of bottom-ranked weights are concentrated in the top layers, returning higher pruning rates as shown in Fig.\,\ref{visualization}(c)\footnote{This phenomenon can also be found in other networks.}.

On the other hand, more FLOPs are typically consumed in the bottom convolutional layers due to the larger input feature maps. Simply considering the magnitude of weights as the importance estimate fails to accelerate inference after pruning. Moreover, different from the weight pruning that simply zeros out certain weight elements but does not change the network structure, filter pruning is more sensitive to the network structure since the whole filters are removed. Thus, it will make the pruned network unstable if a large portion of filters are removed from the top layers. From this view, the per-layer sparsity should be well balanced while also considering per-layer computation.

To this end, we propose to retain the pruned network structure formulation in Eq.\,(\ref{pruning_rate}) for its easy implementation, while redefining a computation-aware importance estimate for each individual weight $(\mathbf{k}_i^j)_q$ as
\begin{equation}\label{normalization}
(\mathbf{\theta}_i^j)_q = \frac{|(\mathbf{k}_i^j)_q|}{(\#FLOPs_i)^{\lambda}},
\end{equation}
where $\#FLOPs_i$ returns the FLOPs count in the $i$-th layer and $\lambda \ge 0$ is a hyper-parameter shared across the network.

It is easy to see that Eq.\,(\ref{normalization}) is a generalization of Eq.\,(\ref{magnitude}). By setting $\lambda = 0$, it degenerates to the magnitude-based importance measurement of Eq.\,(\ref{magnitude}) widely used in~\cite{han2015learning,zhu2018to,gal2019the,mostafa2019parameter,frankle2019lottery}. With a fixed $\lambda$, smaller-magnitude weights with more computation consumption lead to less importance estimates, and then tend to be removed. Therefore, it can well tackle the ``long-tail'' pruning problem arising from Eq.\,(\ref{magnitude}). Moreover, our pruned network structure using Eq.\,(\ref{sparse}) considers the cross-layer ranking of pre-trained weights. As validated in Sec.\,\ref{experiment}, a better performance can be obtained since the global relationship is considered. Besides, it can be easily implemented without any complex learning requirement, which differs our method from existing search-based works~\cite{liu2019metapruning,lin2020channel}.

\subsection{$k$-Reciprocal Nearest Filters}

Given the $i$-th-layer pruning rate $p_i$ determined by Eq.\,(\ref{pruning_rate}) based on the importance estimate in Eq.\,(\ref{normalization}), we have the number of preserved filters: $\bar{n}_i = \big\lceil (1 - p_i) \cdot n_i \big\rfloor$. The next step in filter pruning lies in finding $\bar{n}_i$ most important filters in the pre-trained model, which would then be transferred to the pruned network structure for the follow-up fine-tuning. Our method for selecting important filter weights lies in measuring the collective importance of a filter subset with size of $\bar{n}_i$ rather than simply considering the individual filter importance in most previous methods~\cite{li2017pruning, he2017channel,yu2018nisp,zhao2019variational,lin2020hrank}. Our insight is that  since the filters in each layer work collectively to achieve a desired outcome, we should consider the collective importance of all candidate filters. To this end, as outlined in Fig.\,\ref{rnf}, we propose a recommendation-based filter selection framework where each filter can suggest a group of $k$ filters which have a higher potential to be inherited by the pruned network structure. And then, the final selected filter set is picked up from these groups according to our introduced $k$-reciprocal nearest filters.

To that effect, we first build the similarity matrix $\mathbf{S}_i \in \mathbb{R}^{n_i \times n_i}$ to model the normalized closeness among the $i$-th layer pre-trained filters $\mathbf{K}_i$, whose elements are defined as
\begin{equation}\label{similarity}
\begin{split}
\mathbf{S}_i^{jh} =& \frac{\exp\big(-\mathcal{D}^2(\mathbf{k}_i^j, \mathbf{k}_i^h)\big)}{\sum_{g=1}^{n_i}\exp\big(-\mathcal{D}^2(\mathbf{k}_i^j, \mathbf{k}_i^g)\big)}, \\&  1 \le i \le L, \; 1 \le j, \; h \le n_i,
\end{split}
\end{equation}
where $\mathcal{D}(\cdot , \cdot)$ is a distance function. While other metrics can be used, we simply consider the $\ell_2$-norm in our implementation, which can well reflect the closeness between filter $\mathbf{k}_i^j$ and filter $\mathbf{k}_i^h$ in our empirical observation. Based on the closeness metric, We then further define the closeness rank of filter $\mathbf{k}_i^h$ with respect to filter $\mathbf{k}_i^j$ as follows
\begin{equation}\label{position}
\mathcal{CR}(\mathbf{k}_i^h | \mathbf{k}_i^j) = 1 + \sum_{g=1}^{n_i} \delta(\mathbf{S}_i^{jg} > \mathbf{S}_i^{jh}),
\end{equation}
where $\delta(\cdot)$ is an indicator function, which returns 1 if the input is true, and 0 otherwise.

For the $i$-th layer, each filter $\mathbf{k}_i^j$ would recommend a group of its nearest-neighbor filters in $\mathbf{K}_i$ as the candidates since these filters tend to be much closer to $\mathbf{k}_i^j$. We can then construct a recommendation set with $k$ filters from $\mathbf{k}_i^j$ as 
\begin{equation}\label{neighborhood}
\mathcal{N}_{\mathbf{k}_i^j}^{k} = \{ \mathbf{k}_i^h | \mathcal{CR}(\mathbf{k}_i^h | \mathbf{k}_i^j) \le k, h = 1, 2, ..., n_i \},
\end{equation}
which captures the $k$ nearest neighbors ($k$-NN) of filter $\mathbf{k}_i^j$ in $\mathbf{K}_i$. 
Although the $k$-NN filters of each filter in $\mathbf{K}_i$ form good candidates for selecting filters in  filter pruning, it is highly possible that different filters make different recommendations. Simply choosing one of the recommendation sets is inappropriate since the chosen recommendation may be close to the reference while being far away from others. To solve this, we propose the following $k$-reciprocal nearest filter set
\begin{equation}\label{knn}
\bar{\mathbf{K}}_i = \mathcal{N}_{\mathbf{k}_i^1}^{k} \cap \mathcal{N}_{\mathbf{k}_i^2}^{k} \cap \cdots \cap \mathcal{N}_{\mathbf{k}_i^{n_i}}^{k}.
\end{equation}

As can be seen, the $k$-reciprocal nearest filter set is defined as the intersection of the $k$-NNs of all filters in $\mathbf{K}_i$. It puts a stricter requirement on the final selected filter set that each picked filter $\bar{\mathbf{k}}_i^j \in \bar{\mathbf{K}}_i$ should fall into the $k$-NN of every pre-trained filter rather than a single one. Thus, some of the low-value neighbors, \emph{i.e.}, close to a particular filter but far away from others, can be excluded.

The size of  $\bar{\mathbf{K}}_i$ may be smaller than the target number of preserved filters $\bar{n}_i$, \emph{i.e.}, $|\bar{\mathbf{K}}_i| < \bar{n}_i$. To solve it, as shown in Fig.\,\ref{rnf}, starting with $k = \bar{n}_i$, we increase the value of \textit{k} with a step of 1 until the number of filters in $\bar{\mathbf{K}}_i$ reaches $\bar{n}_i$.

We summarize our pruning steps in Alg.\,\ref{alg1}. As shown, Line 1 -- Line 12 summarize our CLR component for the pruned network structure and Line 13 -- Line 19 outline our RNF part for the filter selection. Both our CLR and RNF are non-learning processes, which significantly reduce the pruning complexity and differentiate our method from existing works.

\begin{algorithm}[!t]
\label{alg1}
\caption{Cross-Layer Ranking \& \textit{k}-Reciprocal Nearest Filters for Pruning Deep Neural Networks}
\LinesNumbered
\KwIn{A pre-trained $L$-layer CNN with kernel $\mathbf{K}$, global pruning rate $p$.}
\KwOut{A compressed representation $\bar{\mathbf{K}}$.}
\For{i = 1 $\rightarrow$ L}{
   \For{j = 1 $\rightarrow$ $n_i$}{
          \ForEach{$(\mathbf{k}_i^j)_q \in \mathbf{k}_i^j$}{
            Calculate the weight importance $({\theta}_i^j)_q$ via Eq.\,(\ref{normalization})\;
          }
   }
}
Conduct a global ranking of weights by $(\mathbf{k}_i^j)_q$\;
Obtain pruning rate $p$ by removing the bottom-ranked weights via Eq.\,(\ref{sparse})\;
\For{i = 1 $\rightarrow$ L}{
   Obtain the per-layer pruning rate $p_i$ via Eq.\,(\ref{pruning_rate})\;
}
\For{i = 1 $\rightarrow$ L}{
   Set $k = \bar{n}_i = \big\lceil (1 - p_i) \cdot n_i \big\rfloor$, $\bar{\mathbf{K}}_i = \{\}$\;
   \While{$| \bar{\mathbf{K}}_i | \neq \bar{n}_i$}{
     Calculate $\bar{\mathbf{K}}_i$ via Eq.\,(\ref{knn})\;
     $k = k + 1$;
     Set $\bar{\mathbf{K}} = \bar{\mathbf{K}} \cap \bar{\mathbf{K}}_i$\; 
   }  
}
Return the compressed representation $\bar{\mathbf{K}}$.
\end{algorithm}

\section{Experiments}\label{experiment}
%

\begin{figure}[!t]
\begin{center}
\includegraphics[height=0.75\linewidth]{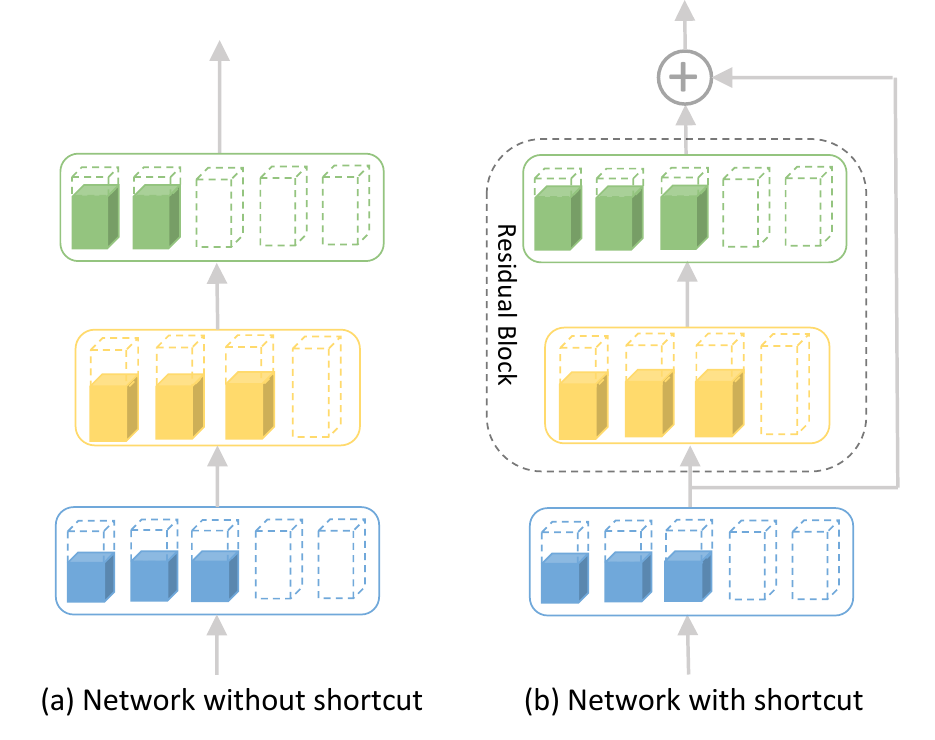}
\end{center}
\caption{\label{pruning}
For networks without shortcut such as VGGNet, GoogleNet, \emph{etc}, we simply remove the filters and their corresponding channels in the next layer (a). For networks with shortcut such as ResNets, following~\cite{ding2019centripetal,liu2019metapruning,lin2020filter,luo2020neural}, we manually reserve the same pruning rate for the input and output layers of each residual block to keep the input and output of the residual block euqal (b). (Best viewed with zooming in)
}
\end{figure}

\subsection{Implementation Settings}\label{implementation}
\subsubsection{Training Details}\label{details}
All our pruned models are fine-tuned via Stochastic Gradient Descent (SGD) optimizer with a momentum of 0.9 and a batch size of 256. On CIFAR-10, we fine-tune each pruned network for 150 epochs with a weight decay of  $5\times$10$^{\text{-3}}$ and an initial learning rate of 0.1, which is decayed to 0.01 and 0.001, respectively, after 50 and 100 epochs. Without specifications, on ImageNet,  we train ResNet-50 for 90 epochs with a weight decay of $1\times$10$^\text{-4}$. The initial learning rate is set to be 0.1 and is divided by 10 every 30 epochs. Without specifications, for all methods, we apply the random crop and horizontal flip to the input images, which are also official operations in Pytorch. To stress, other techniques for image augmentation, such as lightening and color jitter, can be applied to further improve the performance as done in the implementations of~\cite{yu2019slimmable,liu2019metapruning,ding2019approximated}, which however are not considered in this paper. Fig.\,\ref{pruning} displays our pruning strategies for networks with/without the shortcut connections.

\subsubsection{Performance Metrics}\label{performance}
The numbers of FLOPs and parameters and their corresponding pruning rate (denoted as PR) are reported to measure the efficacy of our CLR-RNF and compared methods. The numbers of FLOPs and parameters reflect the computation cost and storage consumptiony. Besides, for CIFAR-10, we report the top-1 accuracy. For ImageNet, both the top-1 and top-5 accuracies are reported.

\subsection{CIFAR-10}\label{results_cifar}
On CIFAR-10, we compare our CLR-RNF with several state-of-the-arts (SOTAs) including~\cite{he2017channel, huang2018data, li2017pruning, lin2020hrank,  lin2020channel, lin2019towards,yu2018nisp,zhao2019variational}. More detailed analyses are provided below.

\textbf{VGGNet}.
We apply our CLR-RNF  to prune the 16-layer VGGNet model, a popular sequential CNN for object detection and semantic segmentation. As shown in Table\,\ref{vggnet_cifar10}, CLR-RNF significantly outperforms the state-of-the-arts for all performance metrics mentioned in Sec. \ref{performance}. Our CLR-RNF can achieve about $20\times$ parameters compression and boost the computation for $4\times$ with even 0.3\% top-1 accuracy improvement, which greatly facilitates the VGGNet model to be deployed on resource-limited devices.

\begin{table}[!t]
\caption{Pruning Results of VGG-16 on CIFAR-10 ($\lambda$ = 0.5). The Numerical Suffix of CLR-RNF Denotes the Global Pruning Rate $p$ in Alg.\,\ref{alg1}.}
\centering
\label{vggnet_cifar10}
\begin{tabular}{cccc}
\toprule
Model                                                 &Top-1 (\%)           &of FLOPs (PR)        &Parameters (PR)\\
\midrule
Baseline~\cite{simonyan2015very}					&93.02 		&314.04M (0.0\%)		&14.73M (0.0\%)\\
SSS~\cite{huang2018data}                             &93.02             &183.13M (41.6\%)    &3.93M (73.8\%) \\
Zhao \emph{et al.}~\cite{zhao2019variational}         &93.18             &190.00M (39.1\%)    &3.92M (73.3\%) \\
GAL-0.05~\cite{lin2019towards}                        &92.03             &189.49M (39.6\%)    &3.36M (77.6\%)  \\
HRank~\cite{lin2020hrank}                            &92.34             &108.61M (65.3\%)    &2.64M (82.1\%)   \\
\textbf{CLR-RNF-0.86}                        &\textbf{93.32} &\textbf{81.31M (74.1\%)} &\textbf{0.74M (95.0\%)} \\
\bottomrule
\end{tabular}
\end{table}

\begin{table}[!t]
\centering
\caption{Pruning Results of GoogLeNet on CIFAR-10 ($\lambda$ = 1). The Numerical Suffix of CLR-RNF Denotes the Global Pruning Rate $p$ in Alg. \,\ref{alg1}.}
\label{googlenet_cifar10}
\begin{tabular}{cccc}
\toprule
Model                                                  &Top-1 (\%)          &FLOPs (PR)       &Parameters (PR)\\
\midrule
Baseline~\cite{szegedy2015going}					&95.03 		&1.53B (0.0\%)		&6.17M (0.0\%)\\
Random                                                 &94.54             &0.96B (36.8\%)    &3.58M (41.8\%)      \\
L1~\cite{li2017pruning}                               &94.54             &1.02B (32.9\%)    &3.51M (42.9\%) \\
GAL-0.05~\cite{lin2019towards}                         &93.93             &0.94B (38.2\%)    &3.12M (49.3\%)  \\
HRank~\cite{lin2020hrank}                             &94.53             &0.69B (54.9\%)    &2.74M (55.4\%)   \\
\textbf{CLR-RNF-0.91}                           &\textbf{94.85}  &\textbf{0.49B (67.9\%)}  &\textbf{2.18M (64.7\%)}  \\
\bottomrule
\end{tabular}
\end{table}

%
\begin{table}[!t]
\centering
\caption{Pruning Results of ResNet-56/110 on CIFAR-10 ($\lambda$ = 10 for ResNet-56 and 5 for ResNet-110). The Digital Numerical Suffix of CLR-RNF Denotes the Global Pruning Rate $p$ in Alg.\,\ref{alg1}.}
\label{resnet_cifar10}
\begin{tabular}{cccc}
\toprule
Model                                                  &Top-1(\%)          &FLOPs (PR)       &Parameters (PR)\\
\midrule
Baseline~\cite{he2016deep}	&93.26 		&126.56M (0.0\%)		&0.85M (0.0\%)\\
L1~\cite{li2017pruning}                               &93.06              &90.90M (27.6\%)   &0.73M (14.1\%) \\
He \emph{et al}.~\cite{he2017channel}                  &90.80              &62.00M (50.6\%)     &-      \\
NISP~\cite{yu2018nisp}     &93.01              &81.00M (35.5\%)     &0.49M (42.4\%)  \\
GAL-0.6~\cite{lin2019towards} &92.90&78.30M (37.6\%)   &0.75M (11.8\%)  \\

FPGM~\cite{he2019filter} &93.26 &59.40M (52.6\%) &- \\
FilterSketch~\cite{lin2020filter}   &93.19  &73.36M (41.5\%) &0.50M (41.2\%)\\
LFPC~\cite{he2020learning} &93.24 &59.10M (52.9\%) &- \\
HRank~\cite{lin2020hrank}   &93.17 &62.72M (50.0\%) &0.49M (42.4\%) \\
SCP~\cite{kang2020operation} &93.23 &61.89M (51.5\%) &0.44M (48.4\%) \\
\textbf{CLR-RNF-0.56}                                &\textbf{93.27}&\textbf{54.00M (57.3\%)}&\textbf{0.38M (55.5\%)}     \\

\midrule
Baseline	~\cite{he2016deep}			&93.57 		&254.99M (0.0\%)		&1.73M (0.0\%)\\
L1~\cite{li2017pruning}                                &93.30             &155.00M (38.7\%)    &1.16M (32.6\%) \\
GAL-0.5~\cite{lin2019towards}                          &92.55             &130.20M (48.5\%)  &0.95M (44.8\%) \\
HRank~\cite{lin2020hrank}                              &93.36             &105.70M (58.2\%)  &0.70M (59.2\%)  \\
LFPC~\cite{he2020learning} & 93.07 &101.00M (60.3\%) & - \\
FilterSketch~\cite{lin2020filter}  &93.44  &92.84M (63.3\%)  &0.69M (59.9\%) \\
\textbf{CLR-RNF-0.69}    &\textbf{93.71} &\textbf{86.80M (66.0\%)} &\textbf{0.53M (69.1\%)}   \\
\bottomrule
\end{tabular}
\end{table}

\textbf{GoogLeNet}.
As shown in Table\,\ref{googlenet_cifar10}, with negligible top-1 accuracy drops (94.85\% for CLR-RNF \emph{vs}. 95.03\% for the baseline), our CLR-RNF can reduce 67.9\% FLOPs and 64.7\% parameters. In comparison with the best state of the art, \emph{i.e.}, HRank, CLR-RNF achieves higher accuracy performance while significantly reducing the numbers and FLOPs and parameters. Thus, CLR-RNF well shows its ability to reduce the redundancy of networks with the multi-branch structure.

\textbf{ResNet}.
We choose to prune ResNet-56 and ResNet-110 to demonstrate the effectiveness of our CLR-RNF for networks with residual blocks. As shown in Table\,\ref{resnet_cifar10}, CLR-RNF takes the lead in both the top-1 accuracy and the FLOPs/parameters compression rates in comparison with the SOTAs. Specifically, for ResNet-56, CLR-RNF reduces the numbers of parameter and FLOPs by 57.3\% and 55.5\%, respectively, without sacrificing the accuracy (93.27\% for CLR-RNF and 93.26\% for the baseline). For ResNet-110, CLR-RNF can reduce 66.0\% FLOPs and 69.1\% parameters while increasing the accuracy performance by 0.14\% (93.71\% for CLR-RNF and 93.57\% for the baseline). Thus, this shows that CLR-RNF can effectively  compress and accelerate networks with residual blocks.

\begin{figure}[!t]
\begin{center}
\includegraphics[height=0.55\linewidth]{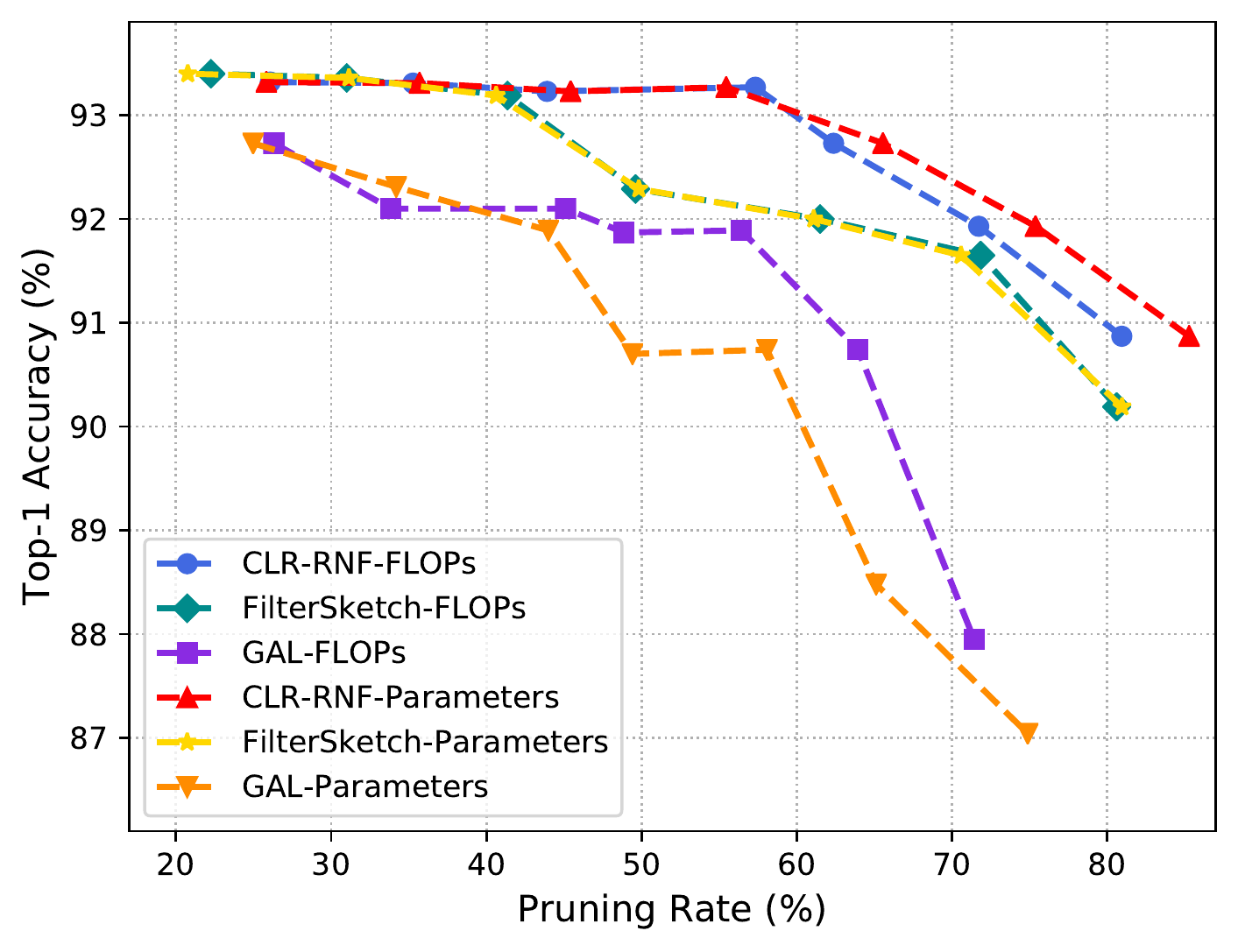}
\end{center}
\caption{\label{vary_rate}Top-1 accuracy comparison between FilterSketch~\cite{lin2020filter}, GAL~\cite{lin2019towards} and our CLR-RNF under similar pruning rates of FLOPs and parameters. The experiments are conducted using ResNet-56.
}
\end{figure}

%
In Fig.\,\ref{vary_rate}, we further compare the top-1 accuracy of the models compressed by GAL~\cite{lin2019towards}, FilterSketch~\cite{lin2020filter} and our CLR-RNF with different pruning rates using ResNet-56. 
As illustrated, GAL performs the worst and suffers severe accuracy drops as the complexity reduction goes deeper. In the case of small pruning rate ($\le 40\%$), our CLR-RNF and FilterSketch achieve similar accuracy performances. However, a large accuracy drop occurs with FilterSketch when the pruning rate is around 50\%, whereas our CLR-RNF can still  well maintain a stable performance. Though the accuracy of CLR-RNF starts a clear drop when the pruning rate is more than 60\%, it still outperforms FilterSketch by a margin and shows an overwhelming gain over GAL, thereby well demonstrating the superiority of CLR-RNF.

\subsection{ImageNet}
We further the results on ImageNet by comparing with several state-of-the-arts~\cite{luo2017thinet,he2017channel,huang2018data,lin2019towards,liu2019metapruning,lin2020hrank,lin2020channel,luo2020autopruner,yang2018netadapt,he2018amc,li2020eagleeye}. 
We compare the accuracy performance under similar FLOPs/parameter reductions or compare the complexity reductions under similar accuracy performance. 

\textbf{ResNet-50}.
Table\,\ref{resnet50_imagenet} shows that CLR-RNF outperforms the other pruning methods in terms of both complexity reduction and accuracy. For example, when setting the global pruning rate $p$ to 0.52, CLR-RNF reduces the pre-trained ResNet-50 to a smaller network with only 0.93B FLOPs and 6.90M parameters. Compared to the search-based ABCPruner-30\% that has 0.94B FLOPs and 7.35M parameters, with more complexity reductions,  CLR-RNF still achieves better performances (71.11\% for CLR-RNF and 70.42\% for ABCPruner in the top-1 accuracy; 90.42\% for CLR-RNF and 89.63\% for ABCPruner in the top-5 accuracy). Similar observations can be found with different values of $p$ such as $0.44$ or $0.20$.

\begin{table}[!t]
\centering
\caption{\label{resnet50_imagenet}Comparison of CLR-RNF ($\lambda$ = 0.4) with Several SOTAs Using ResNet-50 \protect\cite{he2016deep} on ImageNet, Including ThiNet \protect\cite{luo2017thinet}, CP \protect\cite{he2017channel}, SSS \protect\cite{huang2018data}, GAL \protect\cite{lin2019towards}, MetaPruning \protect\cite{liu2019metapruning}, HRank \protect\cite{lin2020hrank}, ABCPruner \protect\cite{lin2020channel}, AutoPruner~\cite{luo2020autopruner} and Slimmable~\cite{yu2019slimmable}. Following~\cite{liu2019metapruning,lin2020channel}, the Numbers of FLOPs and Parameters, and Top-1 and Top-5 Accuracies of the Compressed Models Are Reported. $^{\star}$ Means Our Reproduced Results. $^{\ast}$ Shows Learning Rate with Cosine Scheduler. The Numerical Suffix of CLR-RNF Indicates the Global Pruning Rate $p$ in Alg.\,\ref{alg1}.}
\setlength{\tabcolsep}{0.25em}
\begin{tabular}{c|c|c|c|c}
\hline
Model           &FLOPs   &Parameters   &Top1-acc   &Top5-acc      \\ \hline
Baseline~\cite{{he2016deep}} &4.11B   &25.56M    &76.01\%       &92.96\% \\
ThiNet-30~\cite{luo2017thinet}  & 1.10B  &8.66M    & 68.42\%  &88.30\% \\
MetaPruning-0.50$^{\star}$~\cite{liu2019metapruning} &1.03B  &8.12M &69.92\%  &89.60\%   \\
HRank~\cite{lin2020hrank} &0.98B  &8.27M  &69.10\%  &89.58\%   \\
ABCPruner-30\%~\cite{lin2020channel} &0.94B  &7.35M &70.29\%  &89.63\% \\
\textbf{CLR-RNF-0.52} &\textbf{0.93B} &\textbf{6.90M}  &\textbf{71.11\%}  &\textbf{90.42\%}  \\ \hline
SSS-26~\cite{huang2018data} &2.33B  &15.60M &71.82\%  &90.79\%   \\
GAL-0.5~\cite{lin2019towards}    & 2.33B  &21.20M & 71.95\%   &90.94\%   \\
GAL-0.5-joint~\cite{lin2019towards} & 1.84B   &19.31M &71.80\%  &90.82\%  \\
ThiNet-50~\cite{luo2017thinet}  & 1.71B  &12.38M & 71.01\%  &90.02\%  \\
MetaPruning-0.75$^{\star}$~\cite{liu2019metapruning} & 2.26B  &19.81M  &72.17\%   &90.86\%  \\
HRank~\cite{lin2020hrank} &1.55B  &13.37M &71.98\%  &91.01\% \\
ABCPruner-50\%~\cite{lin2020channel} &1.30B  &9.10M &72.58\%  &90.91\% \\
\textbf{CLR-RNF-0.44} &\textbf{1.23B}  &\textbf{9.00M} &\textbf{72.67\%}  &\textbf{91.09\%} \\ \hline
SSS-32~\cite{huang2018data}  &2.82B &18.60M  &74.18\%   &91.91\%  \\
CP~\cite{he2017channel}   & 2.73B &- & 72.30\%  &90.80\%  \\
MetaPruning-0.85$^{\star}$~\cite{liu2019metapruning} & 2.92B  &19.03M  & 74.49\%  &92.14\%   \\
ABCPruner-100\%~\cite{lin2020channel} &2.56B &18.02M  &74.84\% &92.27\% \\
\textbf{CLR-RNF-0.20} &\textbf{2.45B} &\textbf{16.92M} &\textbf{74.85\%} &\textbf{92.31\%} \\ \hline \hline
AutoPruner$^{\ast}$\cite{luo2020autopruner} &1.39B &12.60M &73.05\% &91.25\%  \\ 
\textbf{CLR-RNF-0.44$^{\ast}$} &\textbf{1.23B} &\textbf{9.00M} &\textbf{73.34\%} &\textbf{91.27\%} \\ \hline 
\end{tabular}
\end{table}

Following the recent advances, \emph{e.g.}, AutoPruner~\cite{luo2020autopruner}, we further apply the learning rate with cosine scheduler, where the initial learning rate is set to 0.1 and the weight decay is set to $4 \times 10^{-5}$. A total of 100 training epochs are applied. As shown in Table\,\ref{resnet50_imagenet}, the superiority of CLR-RNF is evident. With significantly reduced model complexity, our CLR-RNF also achieves the top-1 accuracy of 73.34\%, significantly better than AutoPruner of 73.05\%. 

\begin{table}[!t]
\centering
\caption{\label{efficiency}Comparison of Runtime Complexity on Finding Out the Pruned Network Architecture for CLR-RNF Tested on NVIDIA Tesla V100 GPUs, and CLR-RNF Tested on Intel(R) Xeon(R) CPU E5-2620 v4 @2.10GHz.}
\begin{tabular}{c|c|c|c|c}
\hline
             &ABCPruner    &  GPUs  &\textbf{CLR-RNF} & CPUs\\ \hline
VGGNet-16    &5387.24s      &  1    &1.08s             &1  \\ \hline
GoogLeNet    &26967.65s     &  1    &0.03s             &1  \\ \hline
ResNet-56    &5810.51s      &  1    &0.03s                    &1 \\ \hline
ResNet-110    &10565.27s    &  1    &0.05s                 &1 \\ \hline
ResNet-50    &43534.72s     &2      &2.13s             &1  \\ \hline
\end{tabular}
\end{table}

\textbf{Efficiency of Cross-Layer Ranking}.
As stressed in Sec.\,\ref{introduction}, prevalent methods resort to a series of complex learning steps in the decision of pruned network structure, such as search-based strategies~\cite{liu2019metapruning,lin2020channel}. Our cross-layer ranking lies in its simplicity by reranking the weight importance. Table\,\ref{efficiency} compares the runtime complexity between our cross-layer ranking and ABCPruner~\cite{lin2020channel} employing artificial colony bee as the search algorithm. As can be observed, our cross-layer ranking, with a single CPU implementation, consumes only a few seconds to derive the pruned network structure, whereas it takes several hours or even days with ABCPruner on an NVIDIA Tesla V100 GPU platform. Note that, ABCPruner would consume much more time to derive the pruned models on other lower-computing devices such as NVIDIA 1080 GPUs and CPUs. To analyze, the search-based strategy has to repeatedly apply search operations and measure the quality of each structure by a fitness function, both of which are computationally very expensive. Thus, the efficiency of our cross-layer ranking is evident. 

\subsection{Ablation Study}\label{ablation}
In this section, we show the ablation studies to respectively explore the effectiveness of our cross-layer ranking and $k$-reciprocal nearest filters. All the experimental results are conducted on CIFAR-10 using VGGNet, ResNet-56/110 and GoogleNet, and on ImageNet using ResNet-50.

\begin{table}[!t]
\centering
\caption{\label{ablation_study_arch}Top-1 Accuracy of Pruned VGGNet, ResNet-56/110 and GoogLeNet on CIFAR-10, and ResNet-50 on ImageNet with Pruned Network Structures from: CLR denotes our cross-layer ranking, ABC are from~\cite{lin2020channel}, and Human denotes the human-designated network structure used in~\cite{lin2020hrank}.}
\setlength{\tabcolsep}{0.2em}
\begin{tabular}{cccc}
\toprule
Model            &Top1 (\%) &FLOPs (PR) &Parameters (PR)  \\ \hline
\textbf{VGG-CLR} &\textbf{93.32} &\textbf{81.31M (74.1\%)}  &\textbf{0.74M (95.0\%)} \\
VGG-ABC          & 93.01 & 82.81M (73.7\%) &1.67M (88.7\%)  \\
VGG-Human        & 92.91 & 82.93M (73.6\%) & 1.23M (91.6\%) \\
\midrule
\textbf{ResNet-56-CLR} &\textbf{93.27} &\textbf{54.00M (57.3\%)} &\textbf{0.38M (55.5\%)} \\
ResNet-56-ABC          & 93.13 & 58.54M (54.1\%) & 0.39M (54.2\%) \\
ResNet-56-Human        & 92.97 & 58.01M (54.2\%) & 0.38M (55.5\%) \\
\midrule
\textbf{ResNet-110-CLR} &\textbf{93.71} &\textbf{86.80M (66.0\%)} &\textbf{0.53M (69.1\%)} \\
ResNet-110-ABC          & 93.32 & 89.87M (65.0\%) & 0.56M (67.4\%)  \\
ResNet-110-Human        & 93.27 & 96.49M (62.7\%) & 0.62M (64.4\%)  \\
\midrule
\textbf{GoogLeNet-CLR}  &\textbf{94.85} &\textbf{491.54M (67.9\%)} &\textbf{2.18M (64.7\%)}\\
GoogLeNet-ABC           & 94.47 &513.19M (66.6\%)  &2.46M (60.1\%) \\
GoogLeNet-Human         & 94.01 &520.37M (66.1\%) &2.29M (62.9\%)\\
\midrule
\textbf{ResNet-50-CLR}  &\textbf{71.11} &\textbf{0.93B (77.4\%)} &\textbf{6.90M (73.0\%)}\\ 
ResNet-50-ABC  &70.53 &0.94B (77.1\%) &7.35M (71.3\%)\\
ResNet-50-Human  &69.40 &0.96B (76.7\%) &6.92M (72.9\%)\\
\bottomrule
\end{tabular}
\end{table}

\textbf{Effectiveness of cross-layer ranking}.
For comparisons, we also consider the pruned network structures given by the search-based artificial bee colony~\cite{lin2020channel} and human-designated policy~\cite{lin2020hrank}. All strategies are fed with filter weights from our $k$-reciprocal nearest filter. From Table\,\ref{ablation_study_arch}, besides more complexity reductions, our cross-layer ranking achieves better accuracy performances as well, validating the efficacy of our cross-layer ranking to find a better pruned network structure.

\begin{figure}[!t]
\begin{center}
\includegraphics[height=0.5\linewidth]{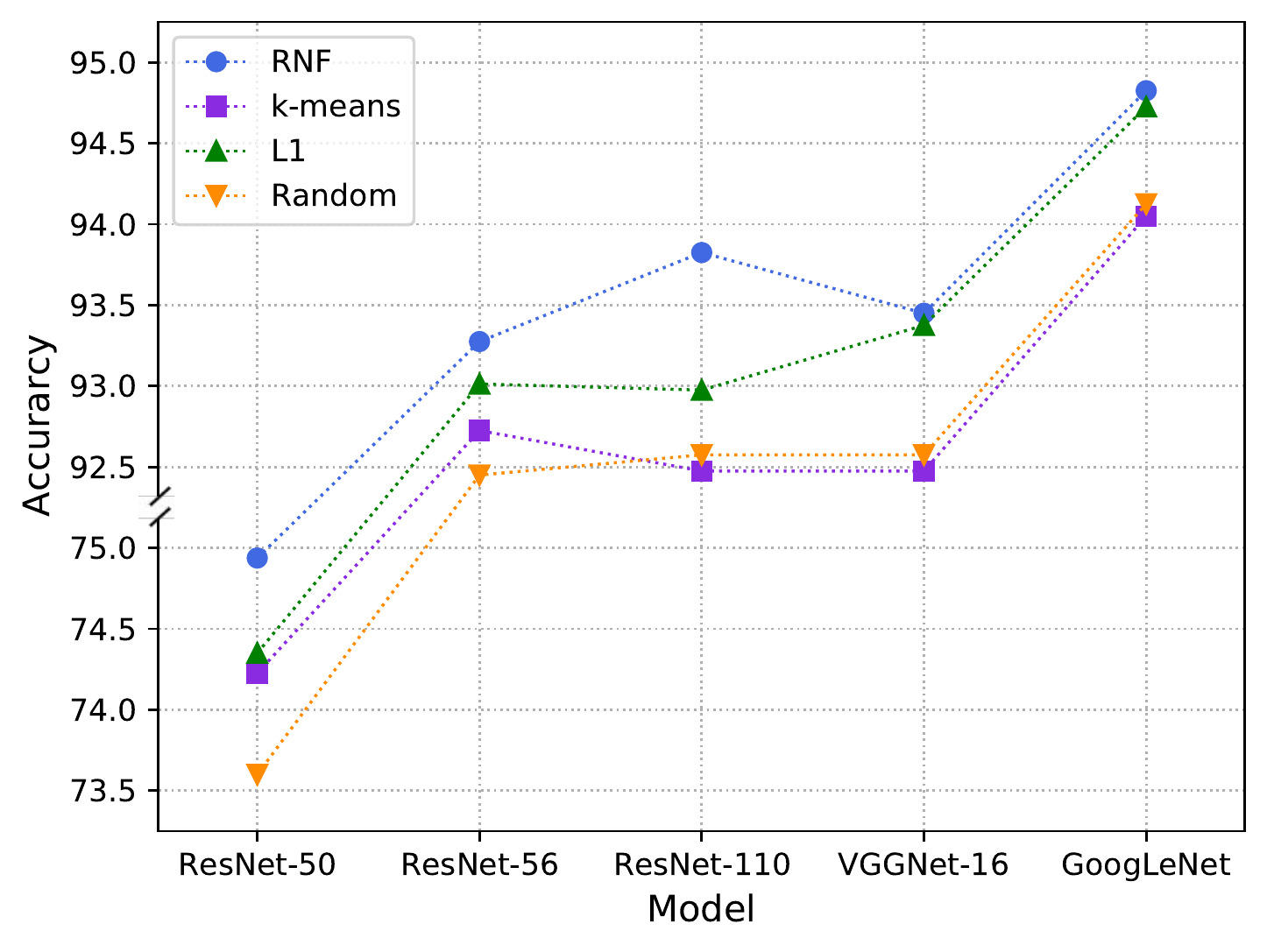}
\end{center}
\vspace{-1.0em}
\caption{\label{ernf}Top-1 accuracy of pruned VGGNet, ResNet-56/110 and GoogLeNet on CIFAR-10, and ResNet-50 on ImageNet. All methods use the same pruned network structures given by our cross-layer ranking.
}
\vspace{-1.0em}
\end{figure}

\textbf{Effectiveness of $k$-reciprocal nearest filter}.
To show the effectiveness of our $k$-reciprocal nearest filter, we further display the performances of different filter selection methods including $k$-means, $\ell_1$-norm, and randomness on top of the same pruned network structures given by our cross-layer ranking. Fig.\,\ref{ernf} shows that our $k$-reciprocal nearest filter outranks the other filter selection scenarios for all networks. This means that our filter selection can recommend a group of more representative filters to constitute the pruned network and rewards a better accuracy performance.

\section{Conclusion}\label{conclusion}
We proposed a novel filter-level network pruning method, called CLR-RNF, involving two non-learning methodologies, cross-layer ranking (CLR) and $k$-reciprocal nearest filter (RNF), that aim to find the optimal pruned network structure and locate a filter subset with better collective importance. To this end, we first revealed the ``long-tail'' pruning problem in the magnitude-based weight pruning and proposed a cross-layer ranking strategy to remove the least important weights ranked by the importance  of individual weights. Furthermore, instead of considering individual filter importance like most previous works, we have devised a recommendation-based filter selection to pick filters with best collective importance. Each filter in the pre-trained model would recommend a group of its closest filters as the potential candidates. Then, the $k$-reciprocal nearest filters that fall into the intersection of different recommendation sets are selected. Extensive experiments on CIFAR-10 and ImageNet demonstrate the efficiency and effectiveness of our new perspective of network pruning.


%



\ifCLASSOPTIONcaptionsoff
  \newpage
\fi



%



\bibliographystyle{IEEEtran}
\bibliography{main}

\begin{thebibliography}{10}
\providecommand{\url}[1]{#1}
\csname url@samestyle\endcsname
\providecommand{\newblock}{\relax}
\providecommand{\bibinfo}[2]{#2}
\providecommand{\BIBentrySTDinterwordspacing}{\spaceskip=0pt\relax}
\providecommand{\BIBentryALTinterwordstretchfactor}{4}
\providecommand{\BIBentryALTinterwordspacing}{\spaceskip=\fontdimen2\font plus
\BIBentryALTinterwordstretchfactor\fontdimen3\font minus
  \fontdimen4\font\relax}
\providecommand{\BIBforeignlanguage}[2]{{%
\expandafter\ifx\csname l@#1\endcsname\relax
\typeout{** WARNING: IEEEtran.bst: No hyphenation pattern has been}%
\typeout{** loaded for the language `#1'. Using the pattern for}%
\typeout{** the default language instead.}%
\else
\language=\csname l@#1\endcsname
\fi
#2}}
\providecommand{\BIBdecl}{\relax}
\BIBdecl

\bibitem{Vanhoucke2011improving}
V.~Vanhoucke, A.~Senior, and M.~Z. Mao, ``Improving the speed of neural
  networks on cpus,'' in \emph{Proc. Adv. Neural Inf. Process. Syst. Workshop},
  2011.

\bibitem{Zhou2017incremental}
A.~Zhou, A.~Yao, Y.~Guo, L.~Xu, and Y.~Chen, ``Incremental network
  quantization: Towards lossless cnns with low-precision weights,'' in
  \emph{Proc. Int. Conf. Learn. Rep.}, 2017.

\bibitem{lin2020rotated}
M.~Lin, R.~Ji, Z.~Xu, B.~Zhang, Y.~Wang, Y.~Wu, F.~Huang, and C.-W. Lin,
  ``Rotated binary neural network,'' in \emph{Proc. Adv. Neural Inf. Process.
  Syst.}, 2020, pp. 7474--7485.

\bibitem{zhang2018shufflenet}
X.~Zhang, X.~Zhou, M.~Lin, and J.~Sun, ``Shufflenet: An extremely efficient
  convolutional neural network for mobile devices,'' in \emph{Proc. IEEE Conf.
  Comput. Vis. Pattern Recognit.}, 2018, pp. 6848--6856.

\bibitem{ma2018shufflenet}
N.~Ma, X.~Zhang, H.-T. Zheng, and J.~Sun, ``Shufflenet v2: Practical guidelines
  for efficient cnn architecture design,'' in \emph{Proc. European Conf.
  Comput. Vis.}, 2018, pp. 116--131.

\bibitem{howard2017mobilenets}
A.~G. Howard, M.~Zhu, B.~Chen, D.~Kalenichenko, W.~Wang, T.~Weyand,
  M.~Andreetto, and H.~Adam, ``Mobilenets: Efficient convolutional neural
  networks for mobile vision applications,'' \emph{arXiv preprint
  arXiv:1704.04861}, 2017.

\bibitem{sandler2018mobilenetv2}
M.~Sandler, A.~Howard, M.~Zhu, A.~Zhmoginov, and L.-C. Chen, ``Mobilenetv2:
  Inverted residuals and linear bottlenecks,'' in \emph{Proc. IEEE Conf.
  Comput. Vis. Pattern Recognit.}, 2018, pp. 4510--4520.

\bibitem{howard2019searching}
A.~Howard, M.~Sandler, G.~Chu, L.-C. Chen, B.~Chen, M.~Tan, W.~Wang, Y.~Zhu,
  R.~Pang, V.~Vasudevan \emph{et~al.}, ``Searching for mobilenetv3,'' in
  \emph{Proc. IEEE Int. Conf. Comput. Vis.}, 2019, pp. 1314--1324.

\bibitem{han2020ghostnet}
K.~Han, Y.~Wang, Q.~Tian, J.~Guo, C.~Xu, and C.~Xu, ``Ghostnet: More features
  from cheap operations,'' in \emph{Proc. IEEE Conf. Comput. Vis. Pattern
  Recognit.}, 2020, pp. 1580--1589.

\bibitem{lin2018holistic}
S.~Lin, R.~Ji, C.~Chen, D.~Tao, and J.~Luo, ``Holistic cnn compression via
  low-rank decomposition with knowledge transfer,'' \emph{IEEE Trans. Pattern
  Anal. Mach. Intell.}, vol.~41, no.~12, pp. 2889--2905, 2018.

\bibitem{hayashi2019exploring}
K.~Hayashi, T.~Yamaguchi, Y.~Sugawara, and S.-i. Maeda, ``Exploring unexplored
  tensor network decompositions for convolutional neural networks,'' in
  \emph{Proc. Adv. Neural Inf. Process. Syst.}, 2019, pp. 5552--5562.

\bibitem{han2015learning}
S.~Han, J.~Pool, J.~Tran, and W.~Dally, ``Learning both weights and connections
  for efficient neural network,'' in \emph{Proc. Adv. Neural Inf. Process.
  Syst.}, 2015, pp. 1135--1143.

\bibitem{frankle2019lottery}
J.~Frankle and M.~Carbin, ``The lottery ticket hypothesis: Finding sparse,
  trainable neural networks,'' in \emph{Proc. Int. Conf. Learn. Rep.}, 2019.

\bibitem{zhou2021learning}
A.~Zhou, Y.~Ma, J.~Zhu, J.~Liu, Z.~Zhang, K.~Yuan, W.~Sun, and H.~Li,
  ``Learning n:m fine-grained structured sparse neural networks from scratch,''
  in \emph{Proc. International Conf. on Learn. Represent.}, 2021.

\bibitem{choquette2021nvidia}
J.~Choquette, W.~Gandhi, O.~Giroux, N.~Stam, and R.~Krashinsky, ``Nvidia a100
  tensor core gpu: Performance and innovation,'' \emph{IEEE Micro}, vol.~41,
  pp. 29--35, 2021.

\bibitem{liu2020autocompress}
N.~Liu, X.~Ma, Z.~Xu, Y.~Wang, J.~Tang, and J.~Ye, ``Autocompress: An automatic
  dnn structured pruning framework for ultra-high compression rates,'' in
  \emph{Proc. AAAI Conf. Artif. Intelli.}, vol.~34, no.~04, 2020, pp.
  4876--4883.

\bibitem{zhang2018structadmm}
T.~Zhang, S.~Ye, K.~Zhang, X.~Ma, N.~Liu, L.~Zhang, J.~Tang, K.~Ma, X.~Lin,
  M.~Fardad \emph{et~al.}, ``Structadmm: A systematic, high-efficiency
  framework of structured weight pruning for dnns,'' \emph{arXiv preprint
  arXiv:1807.11091}, 2018.

\bibitem{mao2017exploring}
H.~Mao, S.~Han, J.~Pool, W.~Li, X.~Liu, Y.~Wang, and W.~J. Dally, ``Exploring
  the granularity of sparsity in convolutional neural networks,'' in
  \emph{Proc. IEEE Conf. Comput. Vis. Pattern Recognit.}, 2017, pp. 13--20.

\bibitem{xie2019exploring}
S.~Xie, A.~Kirillov, R.~Girshick, and K.~He, ``Exploring randomly wired neural
  networks for image recognition,'' in \emph{Proc. IEEE International Conf. on
  Comput. Vis. (ICCV)}, 2019, pp. 1284--1293.

\bibitem{elsen2020fast}
E.~Elsen, M.~Dukhan, T.~Gale, and K.~Simonyan, ``Fast sparse convnets,'' in
  \emph{Proc. IEEE Conf. Comput. Vis. Pattern Recognit.}, 2020, pp.
  14\,629--14\,638.

\bibitem{kalchbrenner2018efficient}
N.~Kalchbrenner, E.~Elsen, K.~Simonyan, S.~Noury, N.~Casagrande, E.~Lockhart,
  F.~Stimberg, A.~Oord, S.~Dieleman, and K.~Kavukcuoglu, ``Efficient neural
  audio synthesis,'' in \emph{Proc. Int. Conf. Mach. Learn.}, 2018, pp.
  2410--2419.

\bibitem{luo2018thinet}
J.-H. Luo, H.~Zhang, H.-Y. Zhou, C.-W. Xie, J.~Wu, and W.~Lin, ``Thinet:
  pruning cnn filters for a thinner net,'' \emph{IEEE Trans. Pattern Anal.
  Mach. Intell.}, vol.~41, no.~10, pp. 2525--2538, 2018.

\bibitem{ding2019centripetal}
X.~Ding, G.~Ding, Y.~Guo, and J.~Han, ``Centripetal sgd for pruning very deep
  convolutional networks with complicated structure,'' in \emph{Proc. IEEE
  Conf. Comput. Vis. Pattern Recognit.}, 2019, pp. 4943--4953.

\bibitem{he2017channel}
Y.~He, X.~Zhang, and J.~Sun, ``Channel pruning for accelerating very deep
  neural networks,'' in \emph{Proc. IEEE Int. Conf. Comput. Vis.}, 2017, pp.
  1389--1397.

\bibitem{huang2018data}
Z.~Huang and N.~Wang, ``Data-driven sparse structure selection for deep neural
  networks,'' in \emph{European Conf. Comput. Vis.}, 2018, pp. 304--320.

\bibitem{liu2019metapruning}
Z.~Liu, H.~Mu, X.~Zhang, Z.~Guo, X.~Yang, K.-T. Cheng, and J.~Sun,
  ``Metapruning: Meta learning for automatic neural network channel pruning,''
  in \emph{Proc. IEEE Int. Conf. Comput. Vis.}, 2019, pp. 3296--3305.

\bibitem{lin2020hrank}
M.~Lin, R.~Ji, Y.~Wang, Y.~Zhang, B.~Zhang, Y.~Tian, and L.~Shao, ``Hrank:
  Filter pruning using high-rank feature map,'' in \emph{Proc. IEEE Conf.
  Comput. Vis. Pattern Recognit.}, 2020, pp. 1529--1538.

\bibitem{li2020eagleeye}
B.~Li, B.~Wu, J.~Su, and G.~Wang, ``Eagleeye: Fast sub-net evaluation for
  efficient neural network pruning,'' in \emph{Proc. European Conf. Comput.
  Vis.}, 2020, pp. 639--654.

\bibitem{liu2019rethinking}
Z.~Liu, M.~Sun, T.~Zhou, G.~Huang, and T.~Darrell, ``Rethinking the value of
  network pruning,'' in \emph{Proc. Int. Conf. Learn. Rep.}, 2019.

\bibitem{lin2020channel}
M.~Lin, R.~Ji, Y.~Zhang, B.~Zhang, Y.~Wu, and Y.~Tian, ``Channel pruning via
  automatic structure search,'' in \emph{Proc. Int. Joint Conf. Artif.
  Intell.}, 2020, pp. 673--679.

\bibitem{liu2017learning}
Z.~Liu, J.~Li, Z.~Shen, G.~Huang, S.~Yan, and C.~Zhang, ``Learning efficient
  convolutional networks through network slimming,'' in \emph{Proc. IEEE Int.
  Conf. Comput. Vis.}, 2017, pp. 2736--2744.

\bibitem{zhao2019variational}
C.~Zhao, B.~Ni, J.~Zhang, Q.~Zhao, W.~Zhang, and Q.~Tian, ``Variational
  convolutional neural network pruning,'' in \emph{Proc. IEEE Conf. Comput.
  Vis. Pattern Recognit.}, 2019, pp. 2780--2789.

\bibitem{luo2020autopruner}
J.-H. Luo and J.~Wu, ``Autopruner: An end-to-end trainable filter pruning
  method for efficient deep model inference,'' \emph{Pattern Recognit.}, p.
  107461, 2020.

\bibitem{lin2020Dynamic}
T.~Lin, S.~U. Stich, L.~Barba, D.~Dmitriev, and M.~Jaggi, ``Dynamic model
  pruning with feedback,'' in \emph{Proc. Int. Conf. Learn. Rep.}, 2020.

\bibitem{li2017pruning}
H.~Li, A.~Kadav, I.~Durdanovic, H.~Samet, and H.~P. Graf, ``Pruning filters for
  efficient convnets,'' in \emph{Proc. Int. Conf. Learn. Rep.}, 2017.

\bibitem{dong2019network}
X.~Dong and Y.~Yang, ``Network pruning via transformable architecture search,''
  in \emph{Proc. Adv. Neural Inf. Process. Syst.}, 2019, pp. 760--771.

\bibitem{yu2019autoslim}
J.~Yu and T.~Huang, ``Autoslim: Towards one-shot architecture search for
  channel numbers,'' \emph{arXiv preprint arXiv:1903.11728}, 2019.

\bibitem{hu2016network}
H.~Hu, R.~Peng, Y.-W. Tai, and C.-K. Tang, ``Network trimming: A data-driven
  neuron pruning approach towards efficient deep architectures,'' \emph{arXiv
  preprint arXiv:1607.03250}, 2016.

\bibitem{ye2018rethinking}
J.~Ye, X.~Lu, Z.~Lin, and J.~Z. Wang, ``Rethinking the
  smaller-norm-less-informative assumption in channel pruning of convolution
  layers,'' in \emph{Proc. Int. Conf. Learn. Rep.}, 2018.

\bibitem{lecun1990optimal}
Y.~LeCun, J.~S. Denker, and S.~A. Solla, ``Optimal brain damage,'' in
  \emph{Proc. Adv. Neural Inf. Process. Syst.}, 1990, pp. 598--605.

\bibitem{dong2017learning}
X.~Dong, S.~Chen, and S.~Pan, ``Learning to prune deep neural networks via
  layer-wise optimal brain surgeon,'' in \emph{Proc. Adv. Neural Inf. Process.
  Syst.}, 2017, pp. 598--605.

\bibitem{ding2019global}
X.~Ding, X.~Zhou, Y.~Guo, J.~Han, J.~Liu \emph{et~al.}, ``Global sparse
  momentum sgd for pruning very deep neural networks,'' in \emph{Proc. Adv.
  Neural Inf. Process. Syst.}, 2019, pp. 6382--6394.

\bibitem{zhu2018to}
M.~Zhu and S.~Gupta, ``To prune, or not to prune: Exploring the efficacy of
  pruning for model compression,'' in \emph{Proc. Int. Conf. Learn. Rep.},
  2018.

\bibitem{gal2019the}
T.~Gale, E.~Elsen, and S.~Hooker, ``The state of sparsity in deep neural
  networks,'' \emph{arXiv preprint arXiv:1902.09574}, 2019.

\bibitem{mostafa2019parameter}
H.~Mostafa and X.~Wang, ``Parameter efficient training of deep convolutional
  neural networks by dynamic sparse reparameterization,'' in \emph{Proc. Int.
  Conf. Mach. Learn.}, 2019, pp. 4646--4655.

\bibitem{ren2020comprehensive}
P.~Ren, Y.~Xiao, X.~Chang, P.-Y. Huang, Z.~Li, X.~Chen, and X.~Wang, ``A
  comprehensive survey of neural architecture search: Challenges and
  solutions,'' \emph{arXiv preprint arXiv:2006.02903}, 2020.

\bibitem{zoph2017neural}
B.~Zoph and Q.~V. Le, ``Neural architecture search with reinforcement
  learning,'' in \emph{Proc. Int. Conf. Learn. Rep.}, 2017.

\bibitem{real2017large}
E.~Real, S.~Moore, A.~Selle, S.~Saxena, Y.~L. Suematsu, J.~Tan, Q.~Le, and
  A.~Kurakin, ``Large-scale evolution of image classifiers,'' in \emph{Proc.
  Int. Conf. Mach. Learn.}, 2017, pp. 2902--2911.

\bibitem{liu2019darts}
H.~Liu, K.~Simonyan, and Y.~Yang, ``Darts: Differentiable architecture
  search,'' in \emph{Proc. Int. Conf. Learn. Rep.}, 2019.

\bibitem{yu2018nisp}
R.~Yu, A.~Li, C.-F. Chen, J.-H. Lai, V.~I. Morariu, X.~Han, M.~Gao, C.-Y. Lin,
  and L.~S. Davis, ``Nisp: Pruning networks using neuron importance score
  propagation,'' in \emph{Proc. IEEE Conf. Comput. Vis. Pattern Recognit.},
  2018, pp. 9194--9203.

\bibitem{lin2020filter}
M.~Lin, L.~Cao, S.~Li, Q.~Ye, Y.~Tian, J.~Liu, Q.~Tian, and R.~Ji, ``Filter
  sketch for network pruning,'' \emph{IEEE Trans. Neural Netw. Learn. Syst.},
  2021.

\bibitem{luo2020neural}
J.-H. Luo and J.~Wu, ``Neural network pruning with residual-connections and
  limited-data,'' in \emph{Proc. IEEE Conf. Comput. Vis. Pattern Recognit.},
  2020, pp. 1458--1467.

\bibitem{yu2019slimmable}
\emph{Slimmable neural networks}, 2019.

\bibitem{ding2019approximated}
X.~Ding, G.~Ding, Y.~Guo, J.~Han, and C.~Yan, ``Approximated oracle filter
  pruning for destructive cnn width optimization,'' in \emph{Proc. Int. Conf.
  Mach. Learn.}, 2019.

\bibitem{lin2019towards}
S.~Lin, R.~Ji, C.~Yan, B.~Zhang, L.~Cao, Q.~Ye, F.~Huang, and D.~Doermann,
  ``Towards optimal structured cnn pruning via generative adversarial
  learning,'' in \emph{Proc. IEEE Conf. Comput. Vis. Pattern Recognit.}, 2019,
  pp. 2790--2799.

\bibitem{simonyan2015very}
K.~Simonyan and A.~Zisserman, ``Very deep convolutional networks for
  large-scale image recognition,'' in \emph{Proc. Int. Conf. Learn. Rep.},
  2015.

\bibitem{szegedy2015going}
C.~Szegedy, W.~Liu, Y.~Jia, P.~Sermanet, S.~Reed, D.~Anguelov, D.~Erhan,
  V.~Vanhoucke, and A.~Rabinovich, ``Going deeper with convolutions,'' in
  \emph{Proc. IEEE Conf. Comput. Vis. Pattern Recognit.}, 2015, pp. 1--9.

\bibitem{he2016deep}
K.~He, X.~Zhang, S.~Ren, and J.~Sun, ``Deep residual learning for image
  recognition,'' in \emph{Proc. IEEE Conf. Comput. Vis. Pattern Recognit.},
  2016, pp. 770--778.

\bibitem{he2019filter}
Y.~He, P.~Liu, Z.~Wang, Z.~Hu, and Y.~Yang, ``Filter pruning via geometric
  median for deep convolutional neural networks acceleration,'' in \emph{Proc.
  IEEE Conf. Comput. Vis. Pattern Recognit.}, 2019, pp. 4340--4349.

\bibitem{he2020learning}
Y.~He, Y.~Ding, P.~Liu, L.~Zhu, H.~Zhang, and Y.~Yang, ``Learning filter
  pruning criteria for deep convolutional neural networks acceleration,'' in
  \emph{Proc. IEEE Conf. Comput. Vis. Pattern Recognit.}, 2020, pp. 2009--2018.

\bibitem{kang2020operation}
M.~Kang and B.~Han, ``Operation-aware soft channel pruning using differentiable
  masks,'' in \emph{Proc. Int. Conf. Mach. Learn.}, 2020.

\bibitem{luo2017thinet}
J.-H. Luo, J.~Wu, and W.~Lin, ``Thinet: A filter level pruning method for deep
  neural network compression,'' in \emph{Proc. IEEE Int. Conf. Comput. Vis.},
  2017, pp. 5058--5066.

\bibitem{yang2018netadapt}
T.-J. Yang, A.~Howard, B.~Chen, X.~Zhang, A.~Go, M.~Sandler, V.~Sze, and
  H.~Adam, ``Netadapt: Platform-aware neural network adaptation for mobile
  applications,'' in \emph{Proc. European Conf. Comput. Vis.}, 2018, pp.
  285--300.

\bibitem{he2018amc}
Y.~He, J.~Lin, Z.~Liu, H.~Wang, L.-J. Li, and S.~Han, ``Amc: Automl for model
  compression and acceleration on mobile devices,'' in \emph{Proc. European
  Conf. Comput. Vis.}, 2018, pp. 784--800.

\end{thebibliography}

\ifCLASSOPTIONcaptionsoff
  \newpage
\fi

%

%

\begin{IEEEbiography}[{\includegraphics[width=1in,height=1.25in,clip,keepaspectratio]{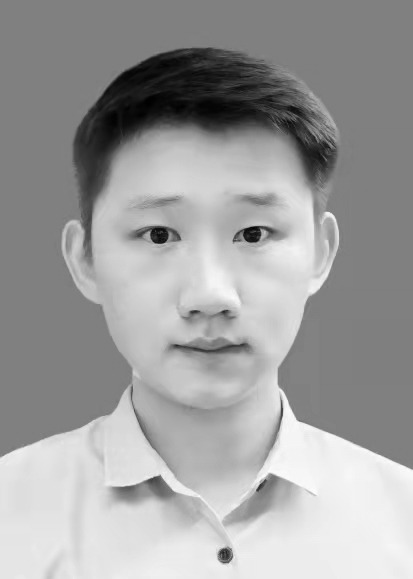}}]{Mingbao Lin} is currently pursuing the Ph.D degree with Xiamen University, China. He has published over ten papers as the first author in international journals and conferences, including IEEE TPAMI, IJCV, IEEE TIP, IEEE TNNLS, IEEE CVPR, NeuriPS, AAAI, IJCAI, ACM MM and so on. His current research interest includes network compression \& acceleration, and information retrieval.
\end{IEEEbiography}

\begin{IEEEbiography}[{\includegraphics[width=1in,height=1.25in,clip,keepaspectratio]{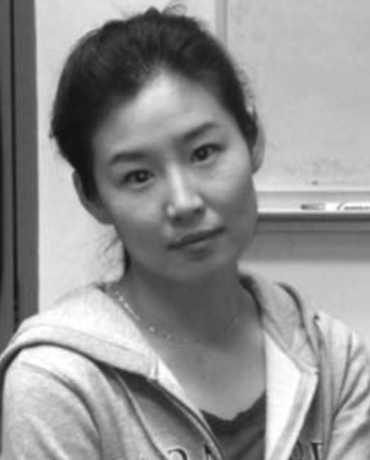}}]{Liujuan Cao}
received the B.S., M.S., and Ph.D degrees from the School of Computer Science and Technology, Harbin Engineering University. She is currently and associate professor at Xiamen University. Her research interests are computer vision and pattern recognition. She has authored over 40 papers in top and major tired journals and conferences, including CVPR, TIP, \emph{etc}. She is the Financial Chair of the IEEE MMSP 2015, the Workshop Chair of the ACM ICIMCS 2016, and the Local Chair of the Visual and Learning Seminar 2017.
\end{IEEEbiography}

\begin{IEEEbiography}[{\includegraphics[width=1in,height=1.25in,clip,keepaspectratio]{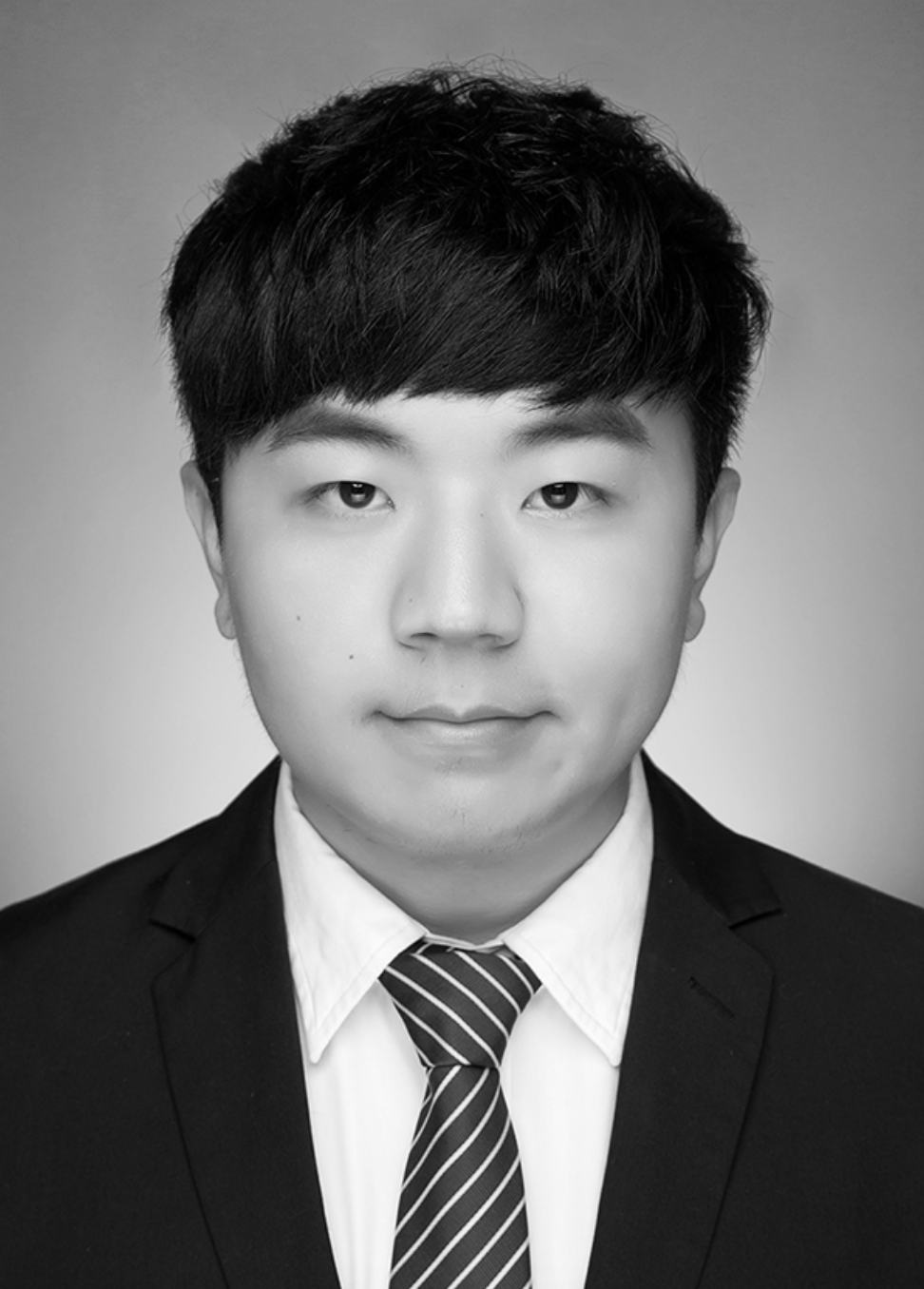}}]{Yuxin Zhang}
is currently pursuing the B.S. degree with Xiamen University, China. His research interests include computer vision, and neural network compression and acceleration.
\end{IEEEbiography}


\begin{IEEEbiography}[{\includegraphics[width=1in,height=1.25in,clip,keepaspectratio]{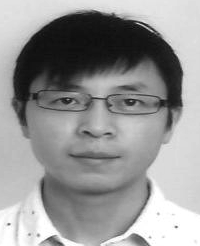}}]{Ling Shao} (Fellow, IEEE) is currently the Executive Vice President and a Provost of the Mohamed bin Zayed University of Artificial Intelligence. He is also the CEO and the Chief Scientist of the Inception Institute of Artificial Intelligence (IIAI), Abu Dhabi, United Arab Emirates. His research interests include computer vision, machine learning, and medical imaging. He is a fellow of IEEE, IAPR, IET, and BCS.
\end{IEEEbiography}

\begin{IEEEbiography}[{\includegraphics[width=1in,height=1.25in,clip,keepaspectratio]{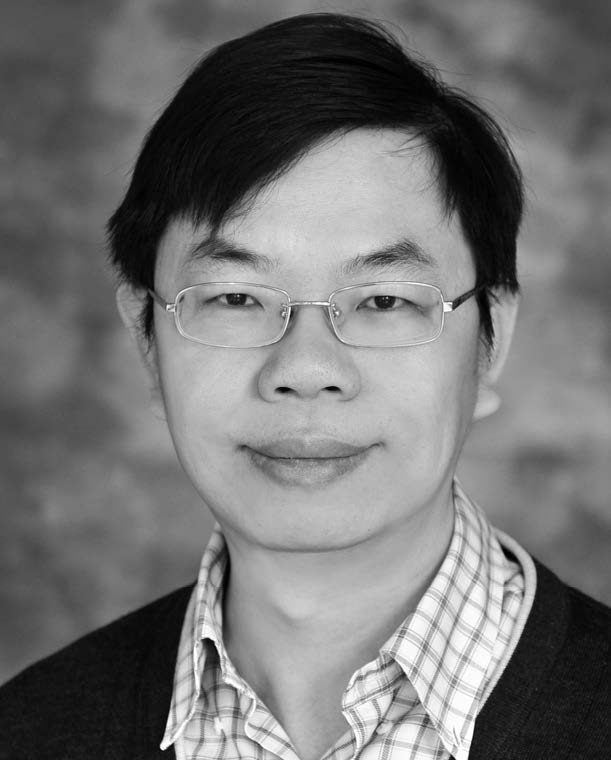}}]{Chia-Wen Lin} (Fellow, IEEE) received the Ph.D degree in electrical engineering from National Tsing Hua University (NTHU), Hsinchu, Taiwan, in 2000.
He is currently Professor with the Department of Electrical Engineering and the Institute of Communications Engineering, NTHU. He is also Deputy Director of the AI Research Center of NTHU. He was with the Department of Computer Science and Information Engineering, National Chung Cheng University, Taiwan, during 2000--2007. Prior to joining academia, he worked for the Information and Communications Research Laboratories, Industrial Technology Research Institute, Hsinchu, Taiwan, during 1992--2000. His research interests include image and video processing, computer vision, and video networking.

Dr. Lin served as  Distinguished Lecturer of IEEE Circuits and Systems Society from 2018 to 2019, a Steering Committee member of \textsc{IEEE Transactions on Multimedia} from 2014 to 2015, and the Chair of the Multimedia Systems and Applications Technical Committee of the IEEE Circuits and Systems Society from 2013 to 2015.  His articles received the Best Paper Award of IEEE VCIP 2015 and the Young Investigator Award of VCIP 2005. He received Outstanding Electrical Professor Award presented by Chinese Institute of Electrical Engineering in 2019, and Young Investigator Award presented by Ministry of Science and Technology, Taiwan, in 2006. He is also the Chair of the Steering Committee of IEEE ICME.  He has served as a Technical Program Co-Chair for IEEE ICME 2010, and a General Co-Chair for IEEE VCIP 2018, and a Technical Program Co-Chair for IEEE ICIP 2019. He has served as an Associate Editor of \textsc{IEEE Transactions on Image Processing}, \textsc{IEEE Transactions on Circuits and Systems for Video Technology}, \textsc{IEEE Transactions on Multimedia}, \textsc{IEEE Multimedia}, and \textit{Journal of Visual Communication and Image Representation}. 
\end{IEEEbiography}

\begin{IEEEbiography}[{\includegraphics[width=1in,height=1.25in,clip,keepaspectratio]{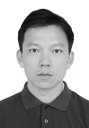}}]{Rongrong Ji}
(Senior Member, IEEE) is a Nanqiang Distinguished Professor at Xiamen University, the Deputy Director of the Office of Science and Technology at Xiamen University, and the Director of Media Analytics and Computing Lab. He was awarded as the National Science Foundation for Excellent Young Scholars (2014), the National Ten Thousand Plan for Young Top Talents (2017), and the National Science Foundation for Distinguished Young Scholars (2020). His research falls in the field of computer vision, multimedia analysis, and machine learning. He has published 50+ papers in ACM/IEEE Transactions, including TPAMI and IJCV, and 100+ full papers on top-tier conferences, such as CVPR and NeurIPS. His publications have got over 10K citations in Google Scholar. He was the recipient of the Best Paper Award of ACM Multimedia 2011. He has served as Area Chairs in top-tier conferences such as CVPR and ACM Multimedia. He is also an Advisory Member for Artificial Intelligence Construction in the Electronic Information Education Committee of the National Ministry of Education.
\end{IEEEbiography}




\end{document}